\documentclass[a4paper,twocolumn,11pt]{quantumarticle}
\pdfoutput=1
\usepackage[utf8]{inputenc}
\usepackage[english]{babel}
\usepackage[T1]{fontenc}
\usepackage{mathtools}
\usepackage{hyperref}
\usepackage{algorithmic}
\usepackage{algorithm}
\usepackage{multirow}
\usepackage{subcaption}

\usepackage{pgfplots}
\pgfplotsset{compat=newest}

\usepackage{tikz}
\usepackage{listofitems}
\usetikzlibrary{arrows.meta}
\usetikzlibrary{decorations.pathreplacing}
\usepackage[subrefformat=parens]{subcaption}
\usepackage{lipsum}
\usepackage{braket}
\usepackage{amsmath,amssymb}

\begin{document}

\title{Quantum Global Variational Learning for Quantum Error Correction}

\author{Shun Ryuzaki}
\affiliation{Computer Science Program, Graduate School of Science and Technology, Meiji University}
\email{ce235047@meiji.ac.jp}
\author{Hideo Mukai}
\affiliation{Computer Science Program, Graduate School of Science and Technology, Meiji University}
\affiliation{Department of Computer Science, School of Science and Technology, Meiji University}
\email{mukai@meiji.ac.jp}
\maketitle

\begin{abstract}
\label{ch:00}
Efficient quantum error correction is essential for the advancement of quantum computing. We propose a quantum neural network with a global structure that reduces the number of unitary matrices required in quantum circuits. This approach resulted in a 97\% reduction in training time and up to a 25\% improvement in the training completion rate, ultimately achieving a 100\% success rate in training while surpassing the error correction performance reported in previous studies. In addition, we demonstrated the enhanced robustness of quantum error correction against internal network noise. Moreover, the fidelity of quantum error correction under internal network noise increased by up to 15\% due to the reduced computational load.
\end{abstract}
\section{Introduction}
\label{ch:01}
One of the critical obstacles in the development of quantum computers is the identification and correction of errors in quantum computation within a reasonable timeframe~\cite{preskill2015lecture}. Information stored in quantum bits (qubits) is highly sensitive and vulnerable to thermal disturbances or interference among qubits. As a result, noise in quantum computation is inevitable, making error correction a matter of paramount importance \cite{rep:qec_begin, rep:qec_gkp}.\par
Quantum error correction requires redundant encoding to recover accurate quantum information without delay. The advancement of error correction methods that are both rapid and efficient is urgently needed, since prolonged correction of a given error increases the likelihood of additional errors occurring \cite{rep:qec_gkp, rep:qec_begin2}.\par
Autonomous quantum error correction using quantum machine learning without measurements has garnered significant attention. Conventional quantum error correction performs quantum operations after introducing ancilla qubits. However, measurement of these ancilla qubits collapses superposition states into classical information, and the correct quantum state is restored via feedback based on this classical outcome. By contrast, quantum error correction with quantum machine learning eliminates such measurements, giving rise to a fully self-contained approach that is inherently more suitable for quantum computers, as it avoids processes that may introduce additional errors.\par
Another important challenge is ensuring convergence during training, which can suffer due to saddle points or local minima in the cost function. It has been observed that the learning process becomes increasingly unstable as the depth and width of the network grow.\par
Our proposed model significantly reduced training time without compromising the accuracy of quantum error correction. Furthermore, the shorter training time mitigates convergence failures caused by saddle points or local minima. In the present study, we also addressed the barren plateau problem, which frequently hinders quantum machine learning and variational quantum algorithms (VQAs). Barren plateaus severely restrict the ability to achieve fast convergence toward optimal solutions, or may even render such convergence impossible, due to unfavorable cost function landscapes. This issue strongly depends on parameter initialisation and network structure, with wider or deeper networks generally yielding less stable training.\par
To address this challenge, we propose a model that reduces the number of parameters. This design substantially lowers the probability of convergence failure caused by saddle points or local minima, primarily due to reduced training time. Our approach also enables the scaling of networks that were previously intractable, thereby improving the overall performance of quantum error correction.\par
\section{Background}
\label{ch:02}
In this section, we provide a brief overview of quantum error correction (QEC) to clarify the context of the results presented in this work. We first elaborate on the quantum error correction codes used to protect quantum information from noise.

Next, we introduce the quantum Hamming bound, an important theoretical notion in QEC. The quantum Hamming bound defines the upper limit of QEC performance according to the code length and the type of errors, serving as a fundamental index for evaluating the capability of different QEC schemes.

\subsection{Quantum error correction code}
The central principle of QEC is the encoding of quantum information using redundant codes. This redundancy enables the recovery of the original information even if part of it is corrupted by noise. For example, a single-qubit state
\[
\ket{\psi} = \alpha \ket{0} + \beta \ket{1}
\]
can be encoded as
\begin{equation}
\label{eq:psi}
\ket{\psi_L} = \alpha \ket{0_L} + \beta \ket{1_L}.
\end{equation}

The encoded quantum state belongs to the code space $\mathcal{H}_L$, spanned by the two logical basis states $\ket{0_L}$ and $\ket{1_L}$. When noise causes the encoded state to leave the code space $\mathcal{H}_L$, this deviation can be detected. Consequently, the types and accuracy of correctable errors are determined by the choice of encoding, in particular the definitions of $\ket{0_L}$ and $\ket{1_L}$.

Noise detection and error inference are neatly described using the stabiliser formalism, which leverages concepts from group theory~\cite{Stabiliser}. Stabiliser operators $\mathcal{S}_i$ are defined as Pauli operators, and the group generated by them forms the stabiliser group $\mathcal{S}$. A stabiliser acts on logical states as
\begin{equation}
\label{eq:stable}
\ket{\psi_L} = \mathcal{S}_i \ket{\psi_L},
\end{equation}
such that the code space $\mathcal{H}_L$ is invariant under this subgroup of the Pauli group. A stabiliser generator is a minimal set of operators sufficient to stabilise the code space. A two-dimensional logical code space encoded by $n$ qubits is stabilised by $(n-1)$ independent stabiliser generators.

QEC via the stabiliser formalism proceeds in two stages: error detection and recovery. In the detection stage, all stabiliser generators are measured, yielding an error syndrome. When the quantum state is error-free, \eqref{eq:stable} holds for all stabilisers. When noise is present, the resulting pattern of syndromes identifies which qubit was affected and by what type of error. The recovery step then applies the corresponding correction operation. For example, if a bit-flip error is detected, the affected qubit is flipped again to restore the original state. Because syndrome analysis involves classical computation, the identification and mapping of syndromes to corrections are typically performed by a classical computer.

When the number of stabiliser generators equals the number of physical qubits, the stabiliser code defines a one-dimensional quantum state. More generally, using $n$ physical qubits to encode a single logical qubit, the code space is defined as a two-dimensional subspace stabilised by the group generated from $(n-1)$ stabiliser generators. Thus, by choosing appropriate stabiliser generators, one can properly define the logical code space $\mathcal{H}_L$.

As a concrete example, consider correcting bit-flip errors in an $n$-qubit repetition code. In this case, the logical states in \eqref{eq:psi} are encoded as
\begin{equation}
    \label{eq:repetition}
    \ket{0_L}=\underbrace{\ket{00\cdots0}}_{\text{$n$ qubits}}, \quad
    \ket{1_L}=\underbrace{\ket{11\cdots1}}_{\text{$n$ qubits}}.
\end{equation}
We then select the following stabiliser generators for generality:
\begin{equation}
    \label{eq:repetition_stabiliser}
    \langle Z_1 Z_n, Z_2 Z_n, \ldots, Z_{n-1} Z_n \rangle = \mathcal{S},
\end{equation}
where $Z_i \ (i = 1, 2, \ldots, n)$ denotes a Pauli-$Z$ operator acting on the $i$-th qubit.

Error correction using this stabiliser construction is straightforward. If a bit-flip error occurs on the $k$-th qubit $(k < n)$, the syndrome associated with $Z_k Z_n$ anticommutes with the error operator $X_k$, leading to
\begin{eqnarray}
    \label{eq:rep_0}
    Z_k Z_n X_k \ket{\psi_L} &=& - X_k Z_k Z_n \ket{\psi_L} \nonumber \\
    &=& - X_k \ket{\psi_L}.
\end{eqnarray}

Thus, the sign change in the syndrome unambiguously reveals the location of the error, since the other stabilisers remain unaffected. However, if the error occurs on the $n$-th qubit, all syndromes become negative due to anticommutation with $X_n$. Additionally, when errors affect qubits other than the $n$-th, syndromes can change back from negative to positive, which coincides with the syndrome pattern of multiple errors across the system. Considering probability distributions, it is most likely that a bit-flip on the $n$-th qubit is responsible when the number of negative syndromes exceeds $\tfrac{n}{2}$. Therefore, correction of the $n$-th qubit is given statistical priority.

Based on these observations, we propose an error correction algorithm that incorporates this prioritization. Using this method, we achieved error correction performance approaching the theoretical upper limit when the error rate was below $0.50$.

\subsection{Quantum Hamming Bound}
\label{subsec:hamming}

In designing efficient quantum error correction codes, it is essential to encode as many logical qubits as possible using a smaller number of physical qubits, while retaining sufficient error correction capability. The quantum Hamming bound provides one of the fundamental criteria by imposing a limit on the number of correctable errors in a given number of physical qubits for a code encoding a given number of logical qubits. Although the quantum Hamming bound applies only to non-degenerate codes, it nevertheless establishes an upper bound on the potential of quantum error correction (QEC) with respect to qubit numbers.

We calculated the minimum number of physical qubits $n$ required to reliably correct errors when encoding $k$ logical qubits redundantly into $n$ physical qubits. Suppose errors occur on $j$ physical qubits. Then the number of possible error locations is given by
\[
\begin{psmallmatrix}
    n\\
    j
\end{psmallmatrix}.
\]
Moreover, because each error may be of Pauli type $X$, $Y$, or $Z$, there are in total $3^j$ possible errors. To ensure reliable correction, each error-affected codeword must correspond to an orthogonal subspace of dimension $2^k$ in order to non-degenerately correct $k$ logical qubits. Since these orthogonal subspaces must be contained within the Hilbert space of dimension $2^n$ defined by the $n$ physical qubits, the following inequality must hold:
\begin{equation}
    \label{eq:ham}
    \sum_{j=0}^t
    \begin{pmatrix}
        n\\
        j
    \end{pmatrix}
    3^j 2^k \leq 2^n.
\end{equation}

This condition defines the quantum Hamming bound, which prohibits the existence of non-degenerate codes that encode $k$ logical qubits into fewer than the required $n$ physical qubits determined by the inequality. In the present study, we primarily focus on bit-flip errors and depolarization noise. As an illustrative example, we derive the minimum number of qubits required when assuming at most a single-qubit error, based on Eq.~\eqref{eq:ham}.

First, consider the case of bit-flip errors only. Since this restricts the possible errors from $\{X,Y,Z\}$ to a single error type, the factor $3^j$ in Eq.~\eqref{eq:ham} does not apply. Thus, the code must satisfy:
\begin{equation}
    \label{eq:ham3}
    2(n+1) \leq 2^n.
\end{equation}
The smallest value of $n$ satisfying this inequality is $n=3$, indicating that three physical qubits are sufficient to correct one bit-flip error within an encoded block of three qubits.

Next, we consider $n$-qubit codes capable of correcting all three types of Pauli errors ($X$, $Y$, and $Z$) on a single physical qubit. Assuming the code has the ability to correct one arbitrary error in $n$ qubits, Eq.~\eqref{eq:ham} becomes:
\begin{equation}
    \label{eq:ham5}
    (3n+1)2 \leq 2^n.
\end{equation}
The smallest solution is $n=5$, implying that no non-degenerate quantum code can exist with fewer than five qubits under this condition.

\section{Related studies}
\label{ch:03}
Theory and application of QEC have been extensively explored through a variety of approaches.

Recently, methods based on quantum machine learning have attracted significant attention. Approaching QEC through quantum machine learning enables error correction tailored to the characteristics of each quantum computing device, while avoiding reliance on classical post-processing. This promises both higher performance and faster execution.

Beer et al.\ proposed the Dissipative Quantum Neural Network (DQNN), a feedforward quantum neural network capable of universal quantum computation~\cite{rep:dqnn_train}. Subsequently, Bondarenko and Feldmann demonstrated the preparation of the GHZ state, a specific quantum state, with high accuracy using a quantum autoencoder (QAE)~\cite{rep:qae_ghz}. Building on these developments, Locher et al.\ achieved quantum error correction of random quantum states using a QAE, thereby exploiting the capability of quantum machine learning to address diverse types of noise~\cite{rep:qae_qec}.

In their study, they introduced a systematic method for determining quantum information encodings, which had previously been designed manually, as well as a technique to reduce training time by utilising the inverse matrix of the encoder within the decoder section.

\subsection{Dissipative Quantum Neural Networks}
We focus in particular on quantum error correction implemented by Dissipative Quantum Neural Networks (DQNNs), a class of feed-forward quantum neural networks. While the DQNN is executed on quantum hardware, parameter optimization is performed on classical hardware. Thus, the overall system constitutes a quantum--classical hybrid architecture~\cite{beer2021training, Sharma_2022}.

The network is composed of multiple neurons, where each neuron corresponds to a qubit that retains quantum information. Considering the mapping $\mathcal{E}_k$ from layer $k-1$ to layer $k$, the collection of channels $\mathcal{Q}$ across the entire network is formed by the concatenation of all interlayer mappings $\mathcal{E}_k$.

\begin{equation}
    \rho_{\mathrm{out}} = \mathcal{Q}(\rho_{\mathrm{in}}) = \mathcal{E}_{\mathrm{out}}(\ldots \mathcal{E}_{3}(\mathcal{E}_{2}(\rho_{\mathrm{in}})))
\end{equation}

In the DQNN, an interlayer mapping $\mathcal{E}_k$ is represented by a unitary matrix that transfers the total quantum information of the previous layer to the neurons of the subsequent layer. The mapping $\mathcal{E}_k$ from layer $k-1$ to layer $k$ is illustrated in Fig.~\ref{fig:dqnn-map}. The formal expression is given as follows:

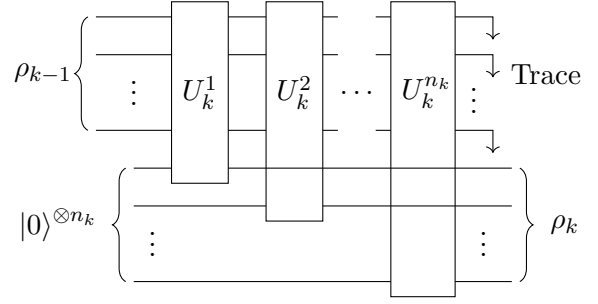
\begin{figure}[!tb]
\centering
\begin{tikzpicture}
\draw(0,10)--(1,10);
\draw(0,9.5)--(1,9.5);
\draw(0.5,9.1)node{$\vdots$};
\draw(0,8.5)--(1,8.5);
\draw [decorate,decoration={brace,amplitude=5pt, raise=3pt},yshift=0pt](0,8.5)--(0,10)node[black,midway,xshift=-0.7cm]{$\rho_{k-1}$};
\draw(1.375,9)node{$U_k^1$};
\draw(1,7.8)--(1.75,7.8)--(1.75,10.2)--(1,10.2)--cycle;
\draw(2.625,9)node{$U_k^2$};
\draw(2.25,7.3)--(3,7.3)--(3,10.2)--(2.25,10.2)--cycle;
\draw(4.375,9)node{$U_k^{n_k}$};
\draw(3.9,6.3)--(4.75,6.3)--(4.75,10.2)--(3.9,10.2)--cycle;
\draw(1.75,10)--(2.25,10);
\draw(1.75,9.5)--(2.25,9.5);
\draw(1.75,8.5)--(2.25,8.5);
\draw(3,10)--(3.2,10);
\draw(3,9.5)--(3.2,9.5);
\draw(3,8.5)--(3.2,8.5);
\draw(3.45,9)node{$\hdots$};
\draw(3.7,10)--(3.9,10);
\draw(3.7,9.5)--(3.9,9.5);
\draw(3.7,8.5)--(3.9,8.5);
\draw(5.35,9.25)node[right]{$\mathrm{Trace}$};
\draw[->](4.75,10)--(5.25,10)->(5.25,9.7);
\draw[->](4.75,9.5)--(5.25,9.5)->(5.25,9.2);
\draw(5,9)node{$\vdots$};
\draw[->](4.75,8.5)--(5.25,8.5)->(5.25,8.2);
\draw [decorate,decoration={brace,amplitude=5pt, raise=3pt},yshift=0pt](0.5,6.5)--(0.5,8.0)node[black,midway,xshift=-1cm]{$\ket{0}^{\otimes n_k}$};
\draw(0.5,8.0)--(1,8.0);
\draw(0.5,7.5)--(2.25,7.5);
\draw(0.75,7.1)node{$\vdots$};
\draw(0.5,6.5)--(3.9,6.5);
\draw(1.75,8.0)--(5.5,8.0);
\draw(3,7.5)--(5.5,7.5);
\draw(5.125,7.1)node{$\vdots$};
\draw(4.75,6.5)--(5.5,6.5);
\draw [decorate,decoration={brace,amplitude=5pt,mirror, raise=3pt},yshift=0pt](5.5,6.5)--(5.5,8)node[black,midway,xshift=0.7cm]{$\rho_k$};
\end{tikzpicture}
\caption{
  Mapping from DQNN layer $k-1$ to layer $k$. The interlayer mapping in the Quantum Autoencoder (QAE) is illustrated. First, the tensor product of the density operator $\rho_{k-1}$ and $n_k$ ancillary states $\ket{0}$ is constructed, representing the expansion of the quantum information from layer $k-1$ together with the initial states $\ket{0}$ assigned to each node in layer $k$. Subsequently, $n_k$ unitary operations $U_k^j$ are applied sequentially to $\rho_{k-1}$ and the ancillary states, where each $U_k^j$ corresponds to an edge connecting layer $k-1$ to the $j$-th node in layer $k$. After all unitary operations are applied, the combined system of $n_{k-1}+n_k$ qubits is reduced to $n_k$ qubits by performing a partial trace over the auxiliary subsystem. The resulting reduced state constitutes the output of the interlayer mapping.
  }
\label{fig:dqnn-map}
\end{figure}

\begin{equation}
    \label{eq:dqnn-map}
    \begin{split}
    \rho_k = &\mathcal{E}_k(\rho_{k-1}) \\
    = &\underset{k-1}{\mathrm{Tr}}[U_k^{n_k}\ldots U_k^1(\rho_{k-1} \otimes \\
    &\ket{0}^{\otimes n_k}\bra{0}^{\otimes n_k})U_k^{1\dag} \ldots U_k^{n_k\dag}]
    \end{split}
\end{equation}

Here, the unitary matrices serve as the training parameters. Each unitary operator represents the connection to a single qubit in the subsequent layer, and sequential application of these operators across layers generates the quantum states of the next layer. In the $k$-th layer, there are $n_k$ unitary operators, and the layer width consists of $(n_{k-1}+1)$ qubits.

For training the process $\mathcal{Q}$, we employ backpropagation in the context of unsupervised learning. The training dataset consists of pairs of input and target quantum states, $(\rho_{\mathrm{in}}, \rho_{\mathrm{targ}})$. The objective of training is to produce the target quantum state $\rho_{\mathrm{targ}}$ from the output state $\rho_{\mathrm{out}} = \mathcal{Q}(\rho_{\mathrm{in}})$, where $\mathcal{Q}$ represents the mapping defined by the network.

In general, the similarity between quantum states is quantified by fidelity. For an output state $\rho_{\mathrm{out}}$ and a target state $\rho_{\mathrm{targ}}$, fidelity is defined as
\begin{eqnarray}
    \label{eq:fid}
    &&\mathcal{F}(\rho_\mathrm{targ}, \rho_\mathrm{out}) \nonumber\\
    &&= \left(\mathrm{Tr}\sqrt{\sqrt{\rho_\mathrm{targ}}\,\rho_\mathrm{out}\,\sqrt{\rho_\mathrm{targ}}}\right)^2 \nonumber\\
    &&= \left(\mathrm{Tr}\sqrt{\ket{\psi_L}\bra{\psi_L}\,\rho_\mathrm{out}\,\ket{\psi_L}\bra{\psi_L}}\right)^2 \nonumber\\
    &&= \left(\sqrt{\bra{\psi_L}\rho_\mathrm{out}\ket{\psi_L}} \,\mathrm{Tr}\sqrt{\ket{\psi_L}\bra{\psi_L}}\right)^2 \nonumber\\
    &&= \bra{\psi_L}\rho_\mathrm{out}\ket{\psi_L}.
\end{eqnarray}

The loss function $\mathcal{L}$ for training in the DQNN is defined as the infidelity, computed using Eq.~\eqref{eq:fid}. The cost function $C$ is then defined as the average of the loss functions evaluated over all training samples. Explicitly, for a target state and the corresponding network output, the loss function $\mathcal{L}$ and the cost function $C$ are given by
\begin{equation}
    \mathcal{L}(\rho_\mathrm{targ}^i,\rho_\mathrm{out}^i) = 1 - \mathcal{F}(\rho_\mathrm{targ}^i,\rho_\mathrm{out}^i),
\end{equation}
\begin{equation}
    C = \frac{1}{N}\sum_{i=1}^{N}\mathcal{L}(\rho_\mathrm{targ}^i,\rho_\mathrm{out}^i).
\end{equation}

To minimise the cost function, the parameters of the unitary operations are updated using classical optimization methods such as gradient descent. Iteratively repeating this process drives the cost function toward its minimum value, thereby improving the performance of the network. Further details of the training procedure are provided in Appendix~\ref{ap:train}.

\subsection{Quantum autoencoders}
We here briefly describe the Quantum Autoencoder (QAE), one of the Dissipative Quantum Neural Networks (DQNNs) extensively employed in the present study~\cite{rep:qae_compression, rep:qae_add, rep:qae_compression2, rep:qae_encode, rep:qae_noise}.

The QAE is a quantum neural network designed for error correction, analogous to the role of autoencoders in classical information processing. The QAE architecture features a hidden layer with fewer qubits than the input and output layers, mirroring its classical counterpart.

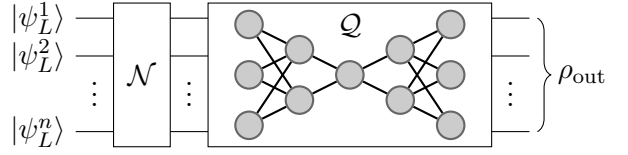
\begin{figure}[!tb]
\centering
\tikzstyle{mynode}=[thick,draw=black!60,fill=black!20,circle,minimum size=5]
\begin{tikzpicture}
  \readlist\Nnod{3,2,1,2,3}
  \foreachitem \N \in \Nnod{
    \foreach \i [evaluate={\x=1.625+\Ncnt*2/3; \y=9.25+(\N/2-\i+0.5)*2/3; \prev=int(\Ncnt-1);}] in {1,...,\N}{
      \node[mynode] (N\Ncnt-\i) at (\x,\y) {};
      \ifnum\Ncnt>1
        \foreach \j in {1,...,\Nnod[\prev]}{
          \draw[thick] (N\prev-\j) -- (N\Ncnt-\i);
        }
      \fi
    }
  }
\draw(0,10)node[left]{$\ket{\psi_{L}^1}$}--(0.5,10);
\draw(0,9.5)node[left]{$\ket{\psi_{L}^2}$}--(0.5,9.5);
\draw(0.25,9.1)node{$\vdots$};
\draw(0,8.5)node[left]{$\ket{\psi_{L}^{n}}$}--(0.5,8.5);
\draw(0.875,9.25)node{$\mathcal{N}$};
\draw(0.5,8.3)--(1.25,8.3)--(1.25,10.2)--(0.5,10.2)--cycle;
\draw(1.25,10)--(1.75,10);
\draw(1.25,9.5)--(1.75,9.5);
\draw(1.5,9.1)node{$\vdots$};
\draw(1.25,8.5)--(1.75,8.5);
\draw(3.625,9.9)node{$\mathcal{Q}$};
\draw(1.75,8.3)--(5.5,8.3)--(5.5,10.2)--(1.75,10.2)--cycle;
\draw(5.5,10)--(6,10);
\draw(5.5,9.5)--(6,9.5);
\draw(5.75,9.1)node{$\vdots$};
\draw(5.5,8.5)--(6,8.5);
\draw [decorate,decoration={brace,amplitude=5pt,mirror, raise=3pt},yshift=0pt](6,8.5)--(6,10)node[black,midway,xshift=0.7cm]{$\rho_\mathrm{out}$};
\end{tikzpicture}
\caption{
Network structure of quantum error correction (QEC) simulating a quantum neural network (QNN) on classical computers.
For the encoded quantum information $\ket{\psi_L}$, the noise network $\mathcal{N}$ introduces noise of a specified type and error rate.
The resulting noisy quantum state is processed by the QNN $\mathcal{Q}$ to yield the corrected, noise-free quantum state $\rho_L$.
The cost function is defined to maximise the fidelity between the target state $\ket{\psi_L}$ and the reconstructed state $\rho_L$.
}
\label{fig:qec_network}
\end{figure}

To realise quantum error correction (QEC) using the QAE, input logical qubits containing noise are processed by the encoder and compressed into a reduced number of qubits in the hidden layer. The decoder then reconstructs the logical qubits, ideally free from noise. The QAE network utilised in this work is depicted in Fig.~\ref{fig:qec_network}.

The encoded quantum state vector $\ket{\psi_L}$ is combined with noise introduced by a noise network, resulting in the noisy input. This state is processed by the feed-forward quantum neural network $\mathcal{Q}$ to produce the output density matrix $\rho_{\mathrm{out}}$.

The noise channel $\mathcal{N}$ generates errors on arbitrary qubits with probability $p$. The QNN $\mathcal{Q}$ receives quantum states with noise as input and outputs noise-reduced quantum states. Accordingly, the training data $(\rho_{\mathrm{in}}, \rho_{\mathrm{targ}})$ are defined as

\begin{equation}
    (\rho_{\mathrm{in}}, \rho_{\mathrm{targ}}) = \left( \mathcal{N}\left( \ket{\psi_L}\bra{\psi_L} \right), \ket{\psi_L}\bra{\psi_L} \right).
\end{equation}

In this study, the number of qubits in the hidden layer is generally set to one for simplicity. An exception is made in Sec.~\ref{sec:noisy_hidden}, where multiple hidden qubits are used to evaluate tolerance to internal network noise. While the number of hidden layers can, in principle, be greater, we limit ourselves to a three-layer architecture in this work. This restriction is motivated by the fact that increasing network depth significantly raises the training cost and reduces tolerance to internal noise.

Overall, using QEC with a QAE, we obtained fidelity values close to the theoretical limits predicted for QEC with stabiliser codes. Furthermore, we established self-exploratory QEC methods capable of autonomously discovering encoding strategies, performing encoding, and operating effectively under internal network noise. Beyond simple QEC, these methods achieved reasonable success in various scenarios.

Nevertheless, one of the remaining challenges is the potential convergence to non-optimal solutions. We observed that saddle points and local minima in the cost function landscape become increasingly pronounced as network size grows. Since enlarging the QEC network is crucial for enhancing its capacity, addressing this problem is of paramount importance.

To overcome this challenge, it is necessary to conduct large-scale training. The solution obtained after convergence heavily depends on the initial parameters of the unitary matrices. Therefore, it is desirable to identify a set of initial conditions that consistently lead to the global optimum while avoiding local minima and saddle points. However, it is evident that finding such initial conditions becomes increasingly difficult as the network expands, both in terms of the number and the size of the unitary matrices.

\section{Quantum global variational learning}
\label{ch:04}
We propose a computationally efficient method to enable the enlargement of QEC networks by mitigating unwanted convergence to barren plateaus, such as saddle points or local minima. To achieve this, we simplify the training process by reducing the number of unitary matrices serving as network parameters. We refer to this approach as Quantum Global Variational Learning (QGVL), as it adopts a single global unitary matrix for interlayer mapping rather than multiple local unitary matrices, as in conventional methods.

\begin{figure}[!tb]
\centering
\begin{tikzpicture}
\draw [decorate,decoration={brace,amplitude=5pt, raise=3pt},yshift=0pt](0,8.5)--(0,10)node[black,midway,xshift=-0.7cm]{$\rho_{k-1}$};
\draw(0,10)--(1,10);
\draw(0,9.5)--(1,9.5);
\draw(0.5,9.1)node{$\vdots$};
\draw(0,8.5)--(1,8.5);
\draw [decorate,decoration={brace,amplitude=5pt, raise=3pt},yshift=0pt](0.5,6.5)--(0.5,8.0)node[black,midway,xshift=-1cm]{$\ket{0}^{\otimes n_k}$};
\draw(0.5,8.0)--(1,8.0);
\draw(0.5,7.5)--(1,7.5);
\draw(0.75,7.1)node{$\vdots$};
\draw(0.5,6.5)--(1,6.5);
\draw(1.5,8)node{$U_k$};
\draw(1,6.3)--(2,6.3)--(2,10.2)--(1,10.2)--cycle;
\draw(2.6,9.25)node[right]{$\mathrm{Trace}$};
\draw[->](2,10)--(2.5,10)->(2.5,9.7);
\draw[->](2,9.5)--(2.5,9.5)->(2.5,9.2);
\draw(2.25,9.1)node{$\vdots$};
\draw[->](2,8.5)--(2.5,8.5)->(2.5,8.2);
\draw(2,8.0)--(3,8.0);
\draw(2,7.5)--(3,7.5);
\draw(2.5,7.1)node{$\vdots$};
\draw(2,6.5)--(3,6.5);
\draw [decorate,decoration={brace,amplitude=5pt,mirror, raise=3pt},yshift=0pt](3,6.5)--(3,8)node[black,midway,xshift=0.7cm]{$\rho_k$};
\end{tikzpicture}
\caption{
Interlayer mapping from layer $k-1$ to layer $k$. The Kronecker product is taken between the input quantum state $\rho_{k-1}$ and the initial ancillary states $\ket{0}^{\otimes n_k}$ corresponding to the output qubits. A partial trace is then performed over the qubits in the input layer to obtain the output state.
}
\label{fig:qgvl}
\end{figure}
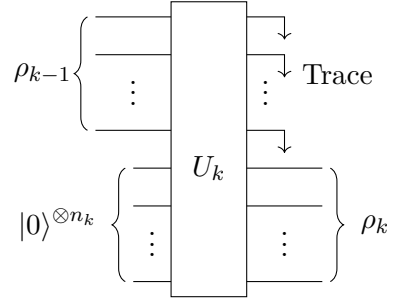

QGVL retains the basic QAE structure shown in Fig.~\ref{fig:qec_network}; however, its most distinctive feature lies in the design of its interlayer mapping, depicted in Fig.~\ref{fig:qgvl}. In the QAE, each unitary matrix connects the entire preceding layer to a single qubit in the subsequent layer, thereby requiring separate unitary operations for each node in the next layer. In contrast, QGVL employs a single, larger unitary matrix that simultaneously incorporates all qubits in both the previous and the next layer. The interlayer mapping in QGVL is formulated as

\begin{equation}
    \label{eq:qgvl-map}
    \begin{split}
    \rho_k &= \mathcal{E}_k(\rho_{k-1}) \\
        &= \mathrm{Tr}_{k-1}\left[ U_k \left( \rho_{k-1} \otimes \ket{0}^{\otimes n_k}\bra{0}^{\otimes n_k} \right) U_k^\dag \right],
    \end{split}
\end{equation}
where $\mathrm{Tr}_{k-1}$ denotes the partial trace over the qubits in layer $k-1$.

The reduction in the number of unitary matrices is significant. For a network with an $m$-$n$-$m$ structure, the QAE requires $(m+n)$ unitary matrices, whereas QGVL requires only two.

We now examine the consequences of employing larger unitary matrices in QGVL compared to QAE. For an $m$-$n$-$m$ architecture, the QAE utilises $n$ matrices of dimension $(m+1)$ and $m$ matrices of dimension $(n+1)$. In contrast, QGVL requires only two matrices, each of dimension $(m+n)$. Although reducing the number of matrices clearly decreases the total number of operations, one might expect that the increased size of the individual matrices in QGVL could raise the computational cost per operation. However, in the QAE, each unitary matrix is extended to match the width of the network layer before execution, resulting in an effective matrix dimension identical to that in QGVL. Consequently, the computational load per matrix operation is equivalent in both approaches.

For further details, including the training procedure, see Appendix~\ref{ap:train}.

For the training of the Quantum Neural Network (QNN), we employed the Rectified Adam (RAdam) optimiser using the norm of comlex gradient matrix as secondary momentum.

\section{Denoising with noiseless channel}
\label{sec:denoised}

First, we compared the performance of quantum error correction (QEC) implemented using the Quantum Autoencoder (QAE) and the Quantum Global Variational Learning (QGVL) approach under the assumption of a noiseless network. As QGVL simplifies the training scheme by reducing the number of parameters relative to QAE, it might be presumed to exhibit lower QEC capability. In this section, we present the results of applying QGVL to two representative quantum error models: X (bit-flip) errors and depolarization errors.

\subsection{Bit flip error correction}
\label{sec:denoised_bitflip}
\subsubsection{3-qubit error correction}
\label{sec:denoised_bitflip_3qubits}

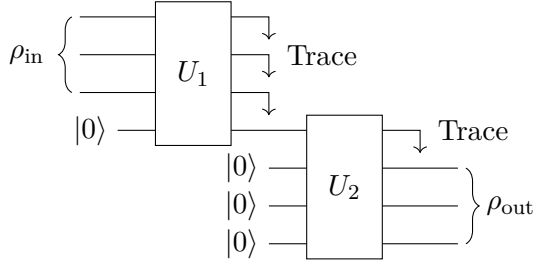
\begin{figure}[!tb]
\centering
\begin{tikzpicture}
\draw(0,10)--(1,10);
\draw(0,9.5)--(1,9.5);
\draw(0,9)--(1,9);
\draw[decorate,decoration={brace,amplitude=5pt, raise=3pt},yshift=0pt](0,9)--(0,10)node[black,midway,xshift=-0.7cm]{$\rho_{\mathrm{in}}$};
\draw(0.5,8.5)node[left]{$\ket{0}$}--(1,8.5);
\draw(1.5,9.25)node{$U_1$};
\draw(1,8.3)--(2,8.3)--(2,10.2)--(1,10.2)--cycle;
\draw(2.6,9.5)node[right]{$\mathrm{Trace}$};
\draw[->](2,10)--(2.5,10)->(2.5,9.7);
\draw[->](2,9.5)--(2.5,9.5)->(2.5,9.2);
\draw[->](2,9)--(2.5,9)->(2.5,8.7);
\draw(2,8.5)--(3,8.5);
\draw(2.5,8)node[left]{$\ket{0}$}--(3,8);
\draw(2.5,7.5)node[left]{$\ket{0}$}--(3,7.5);
\draw(2.5,7)node[left]{$\ket{0}$}--(3,7);
\draw(3.5,7.75)node{$U_2$};
\draw(3,6.8)--(4,6.8)--(4,8.7)--(3,8.7)--cycle;
\draw(4.6,8.5)node[right]{$\mathrm{Trace}$};
\draw[->](4,8.5)--(4.5,8.5)->(4.5,8.2);
\draw(4,8)--(5,8);
\draw(4,7.5)--(5,7.5);
\draw(4,7)--(5,7);
\draw[decorate,decoration={brace,amplitude=5pt,mirror, raise=3pt},yshift=0pt](5,7)--(5,8)node[black,midway,xshift=0.7cm]{$\rho_\mathrm{out}$};
\end{tikzpicture}
\caption{
Quantum Autoencoder (QAE) network architecture consisting of 3-1-3 qubits. The network outputs noise-free quantum information $\rho_\mathrm{out}$ from a 3-qubit input quantum state. The encoder and decoder correspond to the first and second interlayer mappings, respectively, each acting on four qubits. The trainable parameters of the network are the unitary matrices $U_1$ and $U_2$ associated with these interlayer mappings.
}
\label{fig:313}
\end{figure}

Specifically, we describe the implementation of QEC for X errors using the QGVL network on a simple 3-qubit system. The network architecture is defined by a 3-1-3 configuration of qubits, as depicted in Fig.~\ref{fig:313}. The parameters of the proposed network are encapsulated in the unitary matrix $U_1$ for the encoding module and $U_2$ for the decoding module.

For the encoding of quantum states, we employed the 3-qubit repetition code, which defines the code space via the logical codewords $\ket{0_L}$ and $\ket{1_L}$:

\begin{equation}
    \label{eq:3rep}
    \ket{0_L}=\ket{000}, \ket{1_L}=\ket{111}
\end{equation}

The 3-qubit system represents the minimal configuration capable of redundantly encoding single-qubit quantum information and enabling correction of a single error. This property is rigorously guaranteed by Eq.~\eqref{eq:ham}, derived from the quantum Hamming bound. In our experiments, we implemented QEC using both the QGVL framework and the stabiliser code, with the stabiliser code serving as a theoretical benchmark to assess the effectiveness of QGVL. The stabiliser generators for the 3-qubit repetition code are defined as follows:

\begin{equation}
    \label{eq:3stab}
    \begin{split}
        I \otimes Z \otimes Z \\
        Z \otimes Z \otimes I
    \end{split}
\end{equation}

Error detection and correction are performed using the stabiliser group generated by these operators.

We now describe the noise settings used in this study. The parameter $p$ denotes the error rate per qubit. To evaluate the expected value of the logical error rate and determine the threshold for $p$ at which QEC should be applied, we proceed as follows. Without error correction codes, the error rate is $p$, since errors independently affect a single qubit. When error correction codes are deployed, the logical error rate is
\[
p_L = p^3 + 3p^2(1-p),
\]
as neither double- nor triple-qubit errors can be corrected within the code. Thus, under the assumption of a noiseless network, deploying error correction is beneficial as long as $p > p_L$, a condition satisfied for $p < 0.5$. When $p > 0.5$, the likelihood of correction is reversed, since $1-p < 0.5$.

In these experiments, we applied the noise channel $\mathcal{N}$, which induces a Pauli $X$ error on each qubit with probability $p$. The noise operation is defined by

\begin{equation}
    \label{eq:noise}
    \mathcal{N}_p^{bit}(\rho)=(1-p)\rho+pX\rho X
\end{equation}

Training data were organised using the noise network $\mathcal{N}$. First, $10^4$ quantum state vectors $\ket{\psi}$ were randomly generated via the Permuted Congruential Generator (PCG), and then encoded as $\ket{\psi_L}$ using the 3-qubit repetition code. Pairs of quantum states before and after application of the noise network, $(\mathcal{N}(\ket{\psi_L}\bra{\psi_L}), \ket{\psi_L}\bra{\psi_L})$, serve as the input and target states, respectively. Test data for evaluation were also generated.

To compare QEC performance as a function of the error rate $p$, we varied $p$ from $0.00$ to $0.50$ in increments of $0.05$ during both training and evaluation. Model evaluation was performed by comparing the mean fidelity for test data obtained from the trained model to the fidelity achievable by the ideal stabiliser-based QEC code. Fidelity for each data point was evaluated prior to averaging, and comparisons with values based on either physical qubits or the stabiliser code were performed. This approach allows capturing individual correction errors, which would otherwise be masked by averaging. Thus, we analyzed the local error correction capability by inspecting fidelity for each data point. Specifically, we extracted $10^2$ outputs for each data instance and measured the fidelity of the target data at a fixed error rate $p=0.20$.

Additionally, quantum operations involved in error correction were visualised using Quantum Process Tomography (QPT); see Appendix~\ref{ap:qpt} for details. Quantum operations realised by QGVL were compared to those by the stabiliser code and to the absence of any operations. In our experiments, error correction by QGVL at $p=0.20$ was visualised by QPT. For simplicity, QPT input bases were restricted to the Kronecker products of $\ket{0}$ and $\ket{1}$, in accordance with the repetition code. QPT results using alternative bases ($\ket{0}, \ket{1}, \ket{+}, \ket{-}$) are provided in Appendix~\ref{ap:qpt}.

\begin{figure}[!tb]
  \centering
  \includegraphics[width=8cm]{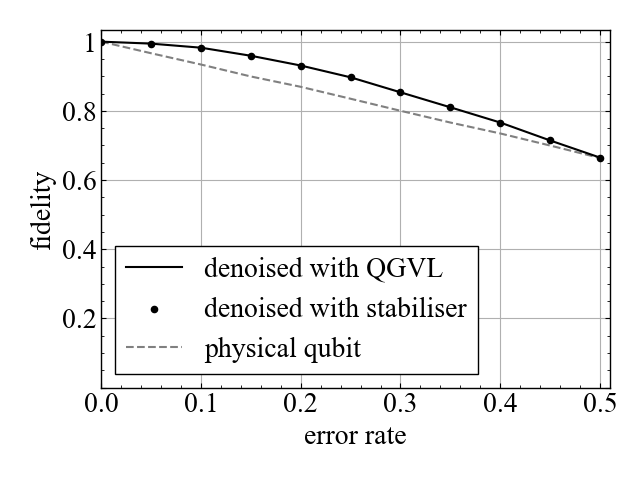}
  \caption{
    Fidelity between target quantum states and error-corrected quantum states obtained using the 3-qubit QGVL (solid line), compared with the fidelity in the case of a single-qubit error among three qubits (dotted line), and the fidelity achieved by algorithmic correction using the stabiliser code (black dots). The target quantum states were generated from $10^4$ randomly prepared quantum states encoded with the repetition code. The input quantum states were produced by introducing bit-flip noise in accordance with the error rates indicated on the x-axis.
  }
  \label{fig:313X}
\end{figure}

The variation in mean fidelity as a function of error rate achieved by QGVL is depicted in Fig.~\ref{fig:313X}. Notably, QEC via QGVL produced results that matched those obtained using the stabiliser code. Mean fidelities exceeding those of the uncorrected physical qubits were observed for $p$ in $(0.00, 0.50)$.

\begin{figure}[!tb]
  \centering
  \includegraphics[width=8cm]{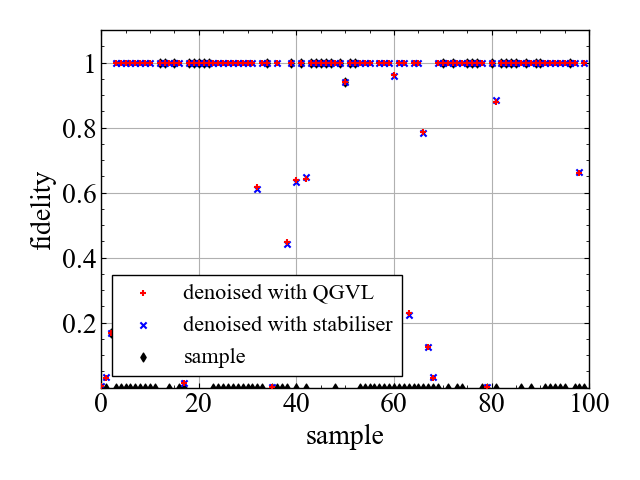}
  \caption{
    Fidelity obtained by the 3-qubit QGVL from $10^2$ test data with a bit-flip error rate of $p=0.1$. Results for physical qubits and stabiliser codes are also shown for comparison. The fidelity of noiseless or fully corrected quantum target states is plotted at 1. In contrast, the fidelity of noisy physical qubits is 0, whereas the fidelities of logical qubits after error correction operations are distributed within the range $(0.0, 1.0)$.
  }
  \label{fig:fidelity}
\end{figure}

Fidelities for individual data points are shown in Fig.~\ref{fig:fidelity}. The fidelities obtained from the QGVL are plotted together with those achieved by the stabiliser code for comparison. While fidelities corresponding to errors in physical qubits are zero, the fidelities of the output states produced by the encoding method are distributed within the range $(0.0, 1.0)$ when errors occur.

\begin{figure*}[!tb]
\centering
    \begin{subfigure}{0.3\linewidth}
        \centering
        \includegraphics[width=\columnwidth]{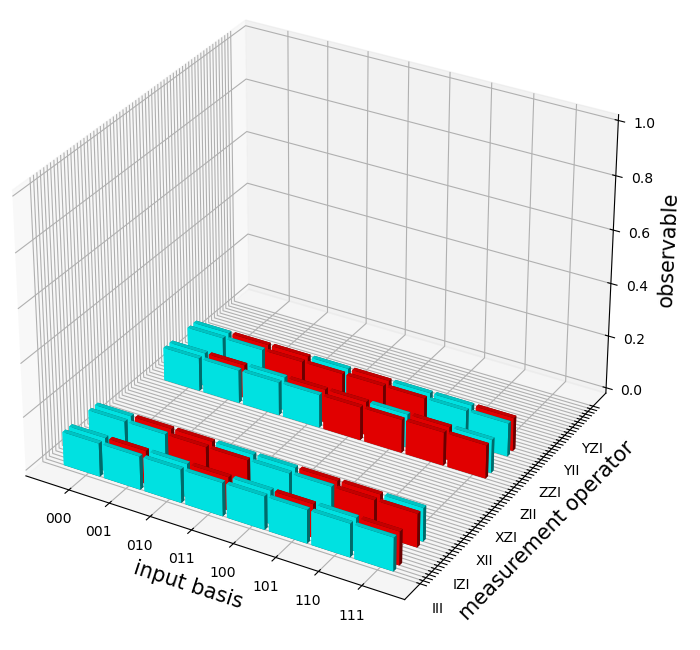}
        \caption{Identity}
        \label{fig:tomography_summary_a}
    \end{subfigure}
    \begin{subfigure}{0.3\linewidth}
        \centering
        \includegraphics[width=\columnwidth]{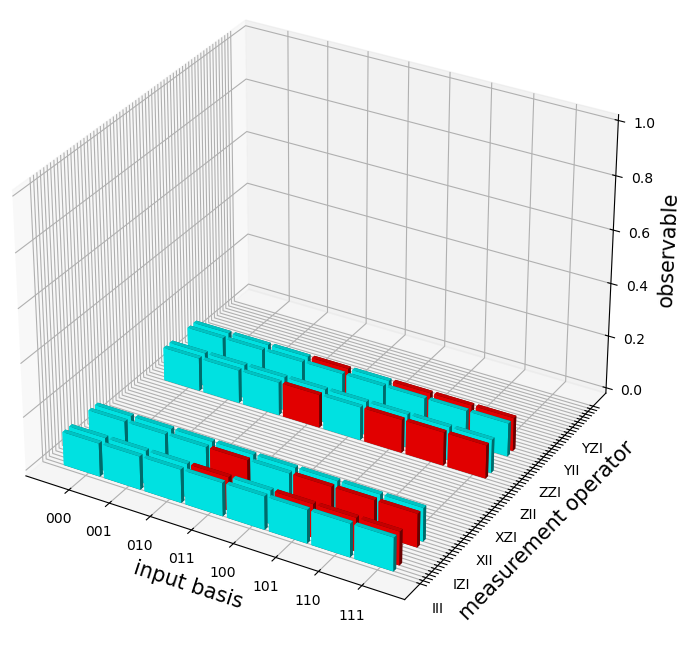}
        \caption{With QGVL}
        \label{fig:tomography_summary_b}
    \end{subfigure}
    \begin{subfigure}{0.3\linewidth}
        \centering
        \includegraphics[width=\columnwidth]{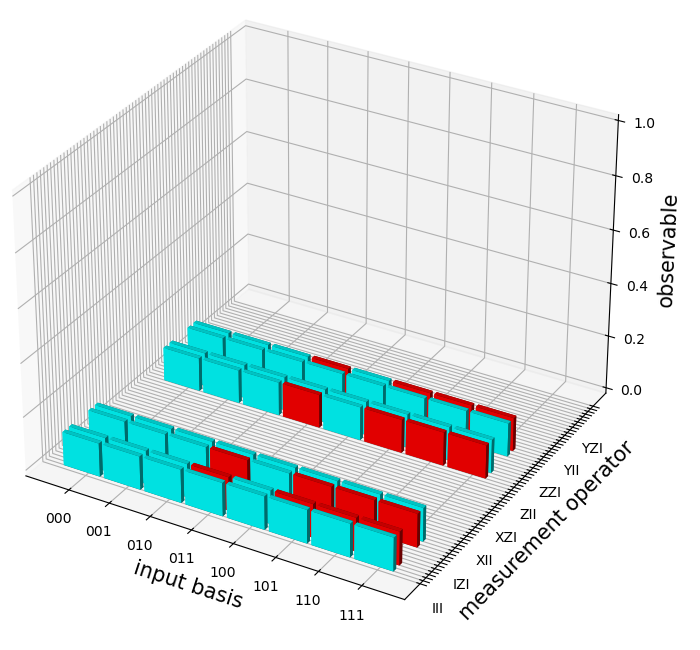}
        \caption{With stabiliser}
        \label{fig:tomography_summary_c}
    \end{subfigure}
    \begin{subfigure}{0.08\linewidth}
        \centering
        \includegraphics[width=\columnwidth]{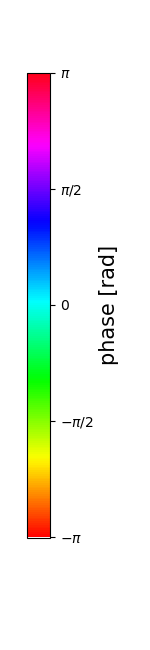}
    \end{subfigure}
    \caption{
    Quantum Process Tomography (QPT) results without any operation (a), using QGVL (b), and employing the stabiliser code (c). The input basis comprised the set $\{\ket{0}, \ket{1}\}^{\otimes 3}$, and the measurement operators were drawn from $\{I, X, Z, Y\}^{\otimes 3}$. Observables correspond to Pauli operators including: $III$, $IIZ$, $IZI$, $IZZ$, $ZII$, $ZIZ$, $ZZI$, and $ZZZ$.
    }
    \label{fig:tomography_summary}
\end{figure*}

In Fig.~\ref{fig:tomography_summary}, the results of quantum operations obtained through Quantum Process Tomography (QPT) are visualised. The observables exhibit positive values when the number of overlaps between the input basis states $\ket{1}$ and the Pauli-$Z$ operator is even, and negative values when it is odd. Moreover, the observation of input bases containing a greater number of $\ket{1}$ states than $\ket{0}$ states corresponds to the expected observables derived from the state $\ket{111}$.

The results of these experiments demonstrate that QGVL achieves quantum error correction with fidelity comparable to that attained by the stabiliser code using the 3-qubit repetition code.

\subsubsection{Comparison with more qubits}
\label{sec:denoised_bitflip_nqubits}

In the previous section, we demonstrated the feasibility of training a 3-qubit quantum error correction (QEC) network. Here, we extend the analysis to multiple-qubit encodings to illustrate the scalability of training larger networks with more qubits, which is the overarching goal in developing robust QEC methods. It is generally expected that increasing the number of qubits improves the accuracy of QEC under idealised conditions, as assumed in prior studies where no noise is present in the qubits constituting the network. Even when internal noise is present, scalability remains critical since the trade-off between error rates and the number of encoding qubits determines the optimal encoding strategy.

In this section, we implement QEC with 5-, 7-, and 9-qubit encodings, extending beyond the 3-qubit case discussed previously. We employ the repetition code appropriate for each qubit number. For training, we generate $10^4$ random quantum states $\ket{\psi_L}$, which are then encoded. The input data are obtained by passing the encoded states through a noise channel, while the target data correspond to the original noise-free states. Test data are generated analogously. The noise error rate $p$ is varied from 0.0 to 4.5 in increments of 0.5. The network’s performance is evaluated by computing the mean fidelity over these datasets and comparing it to that achieved by the stabiliser code across corresponding error rates and qubit numbers.

\begin{figure}[!tb]
  \centering
  \includegraphics[width=8cm]{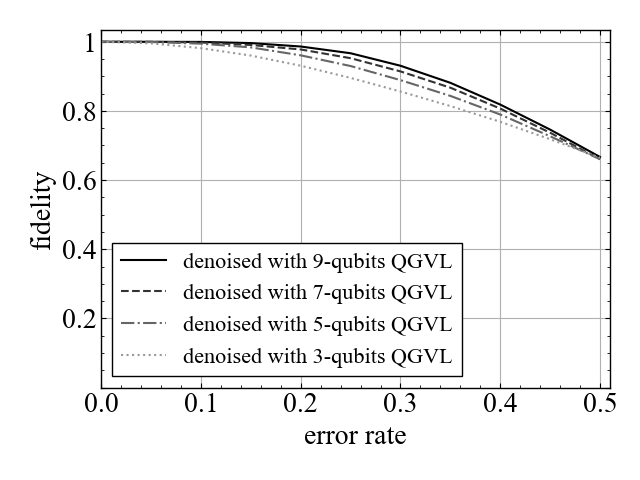}
  \caption{
    Fidelity distributions of error-corrected quantum states by QGVL with 9, 7, 5, and 3 qubits are compared to the fidelities of the corresponding target quantum states. The dataset and noise model are as described above.
    }
  \label{fig:qubitnum}
\end{figure}

The results indicate that for all qubit counts (5, 7, and 9), the mean fidelity achieved by QEC using QGVL matches that of the stabiliser code. Figure~\ref{fig:qubitnum} displays the mean fidelity and associated quantum error rates. As the number of qubits increases from 3 to 9, the mean fidelity improves, demonstrating enhanced QEC accuracy. Furthermore, at error rates of 0 and 0.5, the mean fidelity converges to the same value regardless of the number of qubits, reflecting error-free and uncorrectable regimes, respectively. Under these conditions, the encoding method does not influence the correction accuracy.

\begin{figure}[t]
\centering
\begin{tikzpicture}
\draw(0,10)--(1,10);
\draw(0,9.5)--(1,9.5);
\draw(0.5,9.1)node{$\vdots$};
\draw(0,8.5)--(1,8.5);
\draw [decorate,decoration={brace,amplitude=5pt, raise=3pt},yshift=0pt](0,8.5)--(0,10)node[black,midway,xshift=-0.7cm]{$\rho_{\mathrm{in}}$};
\draw(0.5,8)node[left]{$\ket{0}$}--(1,8);
\draw(1.5,9)node{$U_1$};
\draw(1,7.8)--(2,7.8)--(2,10.2)--(1,10.2)--cycle;
\draw(2.6,9.5)node[right]{$\mathrm{Trace}$};
\draw[->](2,10)--(2.5,10)->(2.5,9.7);
\draw[->](2,9.5)--(2.5,9.5)->(2.5,9.2);
\draw(2.25,9.1)node{$\vdots$};
\draw[->](2,8.5)--(2.5,8.5)->(2.5,8.2);
\draw(2,8)--(3,8);
\draw(2.5,7.5)node[left]{$\ket{0}$}--(3,7.5);
\draw(2.5,7)node[left]{$\ket{0}$}--(3,7);
\draw(2.75,6.6)node{$\vdots$};
\draw(2.5,6)node[left]{$\ket{0}$}--(3,6);
\draw(3.5,7)node{$U_2$};
\draw(3,5.8)--(4,5.8)--(4,8.2)--(3,8.2)--cycle;
\draw(4.6,8.5)node[right]{$\mathrm{Trace}$};
\draw[->](4,8)--(4.5,8)->(4.5,7.7);
\draw(4,7.5)--(5,7.5);
\draw(4,7)--(5,7);
\draw(4.5,6.6)node{$\vdots$};
\draw(4,6)--(5,6);
\draw [decorate,decoration={brace,amplitude=5pt,mirror, raise=3pt},yshift=0pt](5,6)--(5,7.5)node[black,midway,xshift=0.7cm]{$\rho_\mathrm{out}$};
\end{tikzpicture}
\caption{
Multiple-qubit layered QGVL network of the QAE with an $m$-1-$m$ qubit architecture, where the input and output layers each comprise $m$ qubits. Noise-free output quantum information $\rho_\mathrm{out}$ is obtained from the $m$-qubit input quantum information $\rho_\mathrm{in}$. This configuration generalises the network depicted in Fig.~\ref{fig:313}. The network consists of two interlayer mappings, each with a width of $m+1$ qubits: the first serves as the encoder and the second as the decoder. The unitary matrices $U_1$ and $U_2$ associated with these mappings represent the trainable parameters.
}
\label{fig:m1m}
\end{figure}
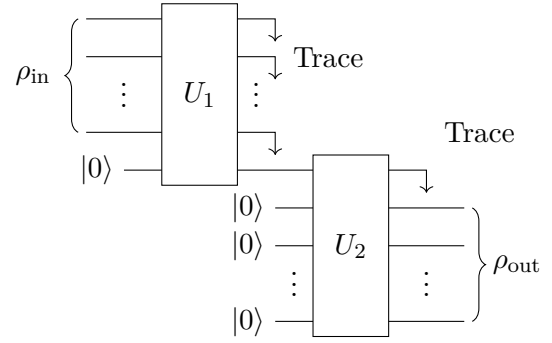

\begin{figure*}[!tb]
\centering
  \begin{minipage}[b]{0.68\columnwidth}
    \centering
    \includegraphics[width=\columnwidth]{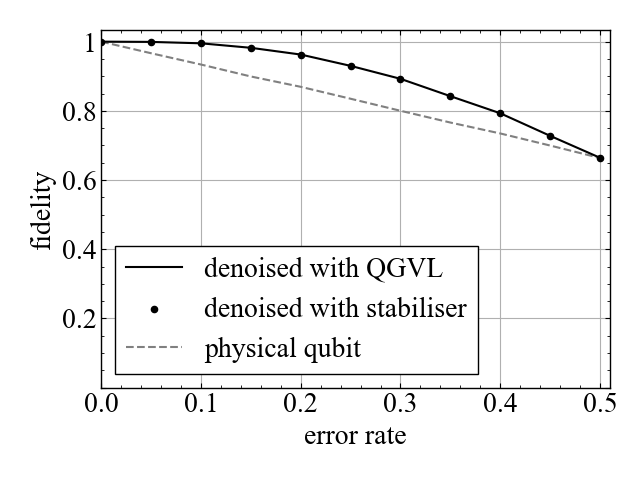}
    \caption*{5-qubit}
  \end{minipage}
  \begin{minipage}[b]{0.68\columnwidth}
    \centering
    \includegraphics[width=\columnwidth]{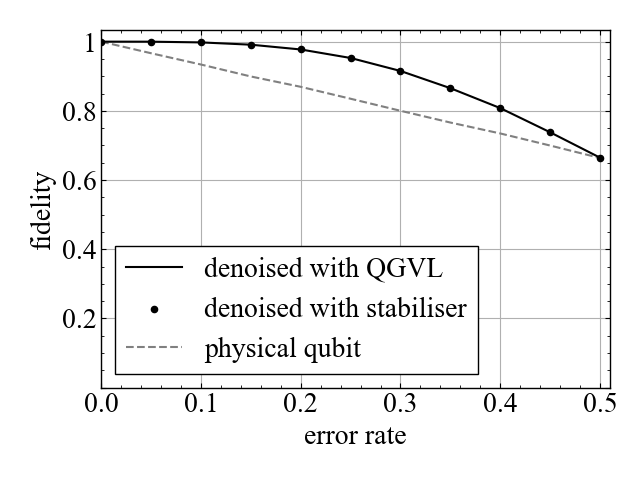}
    \caption*{7-qubit}
  \end{minipage}
  \begin{minipage}[b]{0.68\columnwidth}
    \centering
    \includegraphics[width=\columnwidth]{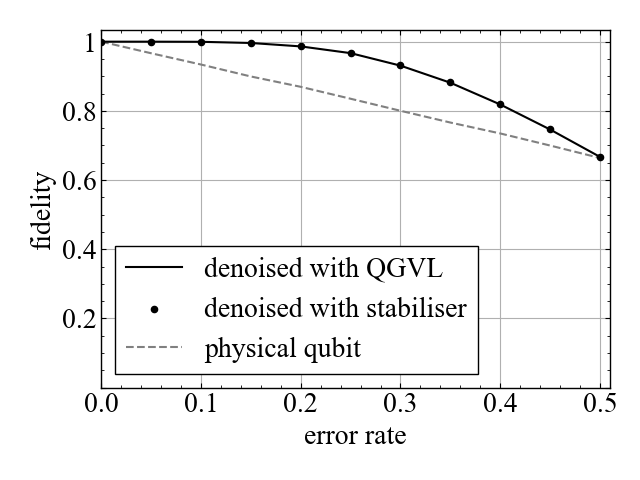}
    \caption*{9-qubit}
  \end{minipage}
  \label{fig:579X}
  \caption{
    Fidelity distributions of quantum states error-corrected by QGVL with 5, 7, and 9 qubits are shown alongside the target quantum states (solid line). For each qubit count, fidelities corresponding to theoretically correctable errors (dotted line) and fidelities achieved through algorithmic quantum error correction using the stabiliser code (black dots) are also plotted. The target quantum states comprise $10^4$ randomly generated quantum states encoded using the repetition code. Input quantum states are generated by applying bit-flip errors to the encoded quantum states, with error rates indicated on the x-axis.
  }
\end{figure*}

\subsection{Depolarising error correction}
In this section, we evaluate the performance of quantum error correction (QEC) under depolarisation noise using the Quantum Global Variational Learning (QGVL) method. Depolarisation errors consist of three types of errors—bit-flip, phase-flip, and bit-phase-flip—occurring with equal probability. This noise model is more complex and realistic than the single-error model considered in the previous section. We begin by implementing QEC with the 5-qubit code, which is the minimal code capable of effectively correcting depolarisation errors. Subsequently, we explore QEC using the Steane and Shor codes, which are important encoding schemes involving larger numbers of qubits.

\subsubsection{5-qubit error correction code}
\label{ch:5-qubit}
We conducted QEC experiments with the smallest QGVL network capable of correcting depolarisation errors. According to the quantum Hamming bound (Eq.~\eqref{eq:ham5}), the minimal number of qubits required for depolarisation-error correction is five. This implies that 1-qubit error can be corrected using a 5-qubit code; thus, we designed a 5-1-5 qubit network corresponding to $m=5$ as shown in Fig.~\ref{fig:m1m}. The trainable parameters of the network are the unitary matrices $U_1$ and $U_2$, acting as the encoder and decoder respectively.

For encoding quantum states, we use the 5-qubit error-correcting code with logical encoding states $\ket{0_L}$ and $\ket{1_L}$ given by

\begin{equation}
\label{eq:depol_encode}
    \begin{split}
    \ket{0_L}=\frac{1}{4}(\ket{00000} + \ket{11000} + \ket{01100} + \ket{00110} \\
    + \ket{00011} + \ket{10001} - \ket{10100} - \ket{01010} \\
    - \ket{00101} - \ket{10010} - \ket{01001} - \ket{11110} \\
    - \ket{01111} - \ket{10111} - \ket{11011} - \ket{11101} )
    \end{split}
\end{equation}
\begin{equation}
    \begin{split}
    \ket{1_L}=&\frac{1}{4}( \ket{11111} + \ket{00111} + \ket{10011} + \ket{11001} \\
    &+ \ket{11100} + \ket{01110} - \ket{01011} - \ket{10101} \\
    &- \ket{11010} - \ket{01101} - \ket{10110} - \ket{00001} \\
    &- \ket{10000} - \ket{01000} - \ket{00100} - \ket{00010} )
    \end{split}
\end{equation}

The QGVL network is trained to acquire QEC capabilities for these encoded quantum states. The stabiliser code is used as the ideal fault-free reference for evaluation. The stabiliser generators for the 5-qubit code are

\begin{equation}
\label{eq:depol_stab}
    \begin{split}
        I \otimes Z \otimes X \otimes X \otimes Z \\
        Z \otimes I \otimes Z \otimes X \otimes X \\
        X \otimes Z \otimes I \otimes Z \otimes X \\
        X \otimes X \otimes Z \otimes I \otimes Z \\
    \end{split}
\end{equation}

and the group generated by these stabilisers enables detection and correction of any single-qubit error.

Next, we describe the noise model. Each qubit is subjected to depolarising errors occurring with probability $p$. The expected logical error rate is computed via the binomial distribution. Without error correction, the error rate naturally equals $p$. Employing the 5-qubit code, errors affecting more than two qubits are uncorrectable. Consequently, the expected logical error rate is
\[
p_L = p^5 + 5 p^4 (1-p) + 10 p^3 (1-p)^2 + 10 p^2 (1-p)^3,
\]
and error suppression occurs provided $p > p_L$. This condition holds for values of $p < 0.13112$ under the assumption of a noiseless network.

In our experiments, the noise channel $\mathcal{N}$ models depolarisation errors with error rate $p$, defined as follows:

\begin{equation}
    \label{eq:noise_depol}
    \begin{split}
    \mathcal{N}_p^{bit}(\rho)=&(1-p)\rho+\frac{p}{3}(X\rho X + Y \rho Y + Z \rho Z)
    \end{split}
\end{equation}

Preparation of training data for depolarisation errors involves a distinct encoding scheme and noise model compared to the bit-flip error case. Randomly generated $10^4$ quantum state vectors $\ket{\psi}$ are encoded as $\ket{\psi_L}$ according to Eq.~\eqref{eq:depol_encode} using the 5-qubit error-correcting code. The pairs of quantum states before and after the application of the noise channel, $(\mathcal{N}(\ket{\psi_L}\bra{\psi_L}), \ket{\psi_L}\bra{\psi_L})$, serve as the input and target states, respectively. Test data are generated and prepared in the same manner.

The error rate $p$ of the noise channel is varied from 0.00 to 0.50 in increments of 0.05. Both the network training process and the corresponding error correction performance are evaluated at each error rate. Performance assessment is based on the mean fidelity computed between the network outputs for test data and their corresponding target states, which is then compared to the fidelity obtained from quantum error correction using the stabiliser code.

\begin{figure}[!tb]
  \centering
  \includegraphics[width=8cm]{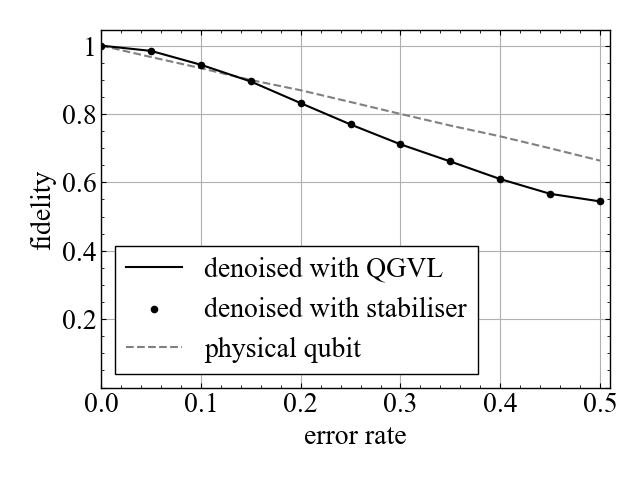}
  \caption{
  Dependence of fidelity on the depolarisation error rate for quantum states corrected by the 5-qubit QGVL network trained with the 5-qubit error-correcting code (solid line). Also shown are fidelities obtained through algorithmic correction using the 5-qubit stabiliser code (black dots) and fidelities of noisy physical single qubits under identical conditions (dotted line). The data comprise $10^4$ randomly generated quantum states. For both QGVL and the stabiliser code, input states are generated by encoding these quantum states with the 5-qubit error-correcting code followed by application of depolarisation errors on each qubit with rate $p$.
  }
  \label{fig:515D}
\end{figure}

The results, presented in Fig.~\ref{fig:515D}, demonstrate that the mean fidelity exhibits a characteristic inflection point and decreases as the quantum error rate increases. The fidelities achieved via algorithmic QEC with the stabiliser code and via QEC using QGVL are largely comparable in magnitude. These findings imply that the QGVL network successfully acquires a QEC strategy capable of correcting depolarisation errors uniformly distributed across qubits.

\subsubsection{Steane code}
As in the case of bit-flip errors, we performed experiments with larger numbers of qubits than in Section~\ref{ch:5-qubit} to evaluate the scalability of QGVL. Here, we assess error correction capabilities using the Steane code, a type of CSS code designed to correct depolarization errors in a 7-qubit system. The Steane code enables error detection and correction via stabiliser generators, which are given by

\begin{equation}
  \label{eq:steane-stabiliser}
  \begin{split}
      I \otimes I \otimes I \otimes X \otimes X \otimes X \otimes X \\
      I \otimes X \otimes X \otimes I \otimes I \otimes X \otimes X \\
      X \otimes I \otimes X \otimes I \otimes X \otimes I \otimes X \\
      I \otimes I \otimes I \otimes Z \otimes Z \otimes Z \otimes Z \\
      I \otimes Z \otimes Z \otimes I \otimes I \otimes Z \otimes Z \\
      Z \otimes I \otimes Z \otimes I \otimes Z \otimes I \otimes Z
  \end{split}
\end{equation}

The first three stabiliser generators correspond to Pauli-$Z$ errors, while the last three address Pauli-$X$ errors. Compared to the [5,1,3] 5-qubit error-correcting code, the Steane code enables correction of both one $X$ and one $Z$ error, as well as individual Pauli errors, at the expense of requiring more qubits. However, not all errors involving two qubits are correctable, and correction is impossible for more than two $X$ and $Z$ errors or for simultaneous errors involving $Y$ operators.

The logical codewords of the Steane code that are associated with the stabiliser generators in Eq.~\eqref{eq:steane-stabiliser} are defined as $\ket{0_L}$ and $\ket{1_L}$:
\begin{equation}
    \label{eq:steane-encoding}
    \begin{split}
    \ket{0_L}=\frac{1}{2\sqrt{2}}(
      &\ket{0000000} + \ket{1101001} \\
    + &\ket{1011010} + \ket{0110011} \\
    + &\ket{0111100} + \ket{1010101} \\
    - &\ket{1100110} - \ket{0001111} )
    \end{split}
\end{equation}
\begin{equation}
    \begin{split}
    \ket{1_L}=\frac{1}{2\sqrt{2}}(
      &\ket{1111111} + \ket{0010110} \\
    + &\ket{0100101} + \ket{1001100} \\
    + &\ket{1000011} + \ket{0101010} \\
    - &\ket{0011001} - \ket{1110000} )
    \end{split}
\end{equation}

As before, the training dataset consists of randomly generated quantum state vectors encoded according to Eq.~\eqref{eq:steane-encoding}. For each encoded state, the noise network~\eqref{eq:noise_depol} introduces depolarization errors with a specified error rate $p$, analogous to the 5-qubit case. Each pair of quantum states after application of the noise network $\mathcal{N}$ and the corresponding original quantum state, $(\mathcal{N}(\psi_L),\psi_L)$, forms the input and target states for training.

For the experiments, the training set comprises $10^4$ randomly generated quantum state vectors. The error rate $p$ of the noise network is varied from $0.00$ to $0.50$ in increments of $0.05$. Network training and evaluation of performance are carried out for each value of $p$. The performance of trained models is assessed by averaging the fidelity between the output of the trained network and the target states on the test data, and comparing this with the theoretical fidelity obtained from QEC using the stabiliser code and that achieved by the QAE.

\begin{figure}[!tb]
  \centering
  \includegraphics[width=8cm]{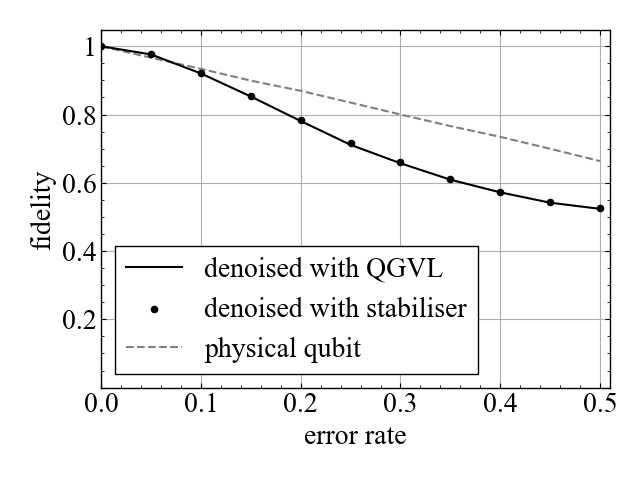}
  \caption{
    Relationship between fidelity and depolarisation error rate for quantum states corrected by the 7-qubit QGVL network trained with the Steane code (solid line), fidelity achieved through algorithmic quantum error correction via the Steane stabiliser code (black dots), and fidelity of physical single-qubit states with depolarisation error under identical conditions (dotted line). For both QGVL and the stabiliser code, input states are prepared by encoding the quantum states using the Steane code and subsequently introducing depolarisation errors with error rate $p$ to each qubit.
  }
  \label{fig:717D}
\end{figure}

The results are presented in Fig.~\ref{fig:717D}. As in the 5-qubit case, the mean fidelity decreases as the quantum error rate increases. Error correction with QGVL achieves a performance level equivalent to algorithmic QEC using the Steane code stabiliser code, demonstrating that QGVL successfully acquires the error correction strategy embodied by the Steane code.

\subsubsection{Shor code}
Finally, we implemented quantum error correction using the Shor code. The Shor code is a 9-qubit, degenerate code that provides protection against depolarisation errors. The logical codewords for the Shor code are $\ket{0_L}$ and $\ket{1_L}$, defined as follows:
\begin{equation}
\label{eq:shor-encode_0}
\ket{0_L}=\frac{1}{2\sqrt{2}}(\ket{000}+\ket{111})^{\otimes 3}
\end{equation}
\begin{equation}
\label{eq:shor-encode_1}
\ket{1_L}=\frac{1}{2\sqrt{2}}(\ket{000}-\ket{111})^{\otimes 3}
\end{equation}
Stabiliser generators for these logical codes are:
\begin{equation}
\label{eq:shor-stabiliser}
    \begin{split}
    &I \otimes I \otimes I \otimes I \otimes I \otimes I \otimes I \otimes Z \otimes Z \\
    &I \otimes I \otimes I \otimes I \otimes I \otimes I \otimes Z \otimes I \otimes Z \\
    &I \otimes I \otimes I \otimes I \otimes Z \otimes Z \otimes I \otimes I \otimes I \\
    &I \otimes I \otimes I \otimes Z \otimes I \otimes Z \otimes I \otimes I \otimes I \\
    &I \otimes Z \otimes Z \otimes I \otimes I \otimes I \otimes I \otimes I \otimes I \\
    &Z \otimes I \otimes Z \otimes I \otimes I \otimes I \otimes I \otimes I \otimes I \\
    &I \otimes I \otimes I \otimes X \otimes X \otimes X \otimes X \otimes X \otimes X \\
    &X \otimes X \otimes X \otimes I \otimes I \otimes I \otimes X \otimes X \otimes X
    \end{split}
\end{equation}
The stabiliser generators for these logical codes are:
\begin{equation}
    \label{eq:shor-error_0}
    \begin{split}
        Z_i\ket{0_L}&=Z_i \frac{1}{\sqrt{2}}(\ket{000}+\ket{111}) \\
        &= \frac{1}{\sqrt{2}}(\ket{000}-\ket{111}) \\
        &= \ket{1_L}
    \end{split}
\end{equation}
\begin{equation}
\label{eq:shor-error_1}
    \begin{split}
        Z_i\ket{1_L}&=Z_i \frac{1}{\sqrt{2}}(\ket{000}-\ket{111}) \\
        &= \frac{1}{\sqrt{2}}(\ket{000}+\ket{111}) \\
        &= \ket{0_L}
    \end{split}
\end{equation}
where $i \in \{0, 1, 2\}$ denotes the specific qubit within a block.

Thus, it is unnecessary to identify which qubit within a block suffered an error, since the change of state is independent of this position. Error correction is executed via a phase flip of any qubit in the block. The Shor code can correct a single Pauli $X$ error within a block and one Pauli $Z$ error across the entire code. Moreover, it can correct a Pauli error that is a superposition of $X$ and $Z$ errors.

Following this coding procedure, both training and test datasets were prepared. Specifically, $5.0 \times 10^4$ random quantum state vectors were encoded via each code. The larger dataset was necessary to encompass all noise patterns resulting from the increased number of qubits and possible error combinations. These states were passed through the noise channel $\mathcal{N}$ to obtain pairs $(\mathcal{N}(\ket{\psi_L}\bra{\psi_L}), \ket{\psi_L}\bra{\psi_L})$ as input and target data. Test sets were generated analogously.

Training and validation of the network were performed with noise error rates $p$ ranging from 0.00 to 0.50 in increments of 0.05. Model validation was conducted by comparing the mean fidelity between the model outputs on test data and the target states, as well as the fidelity achievable by QEC using the stabiliser code. The impact of qubit number on mean fidelity was examined by comparing results for the 5-qubit code, Steane code, and Shor code.

\begin{figure}[!tb]
  \centering
  \includegraphics[width=8cm]{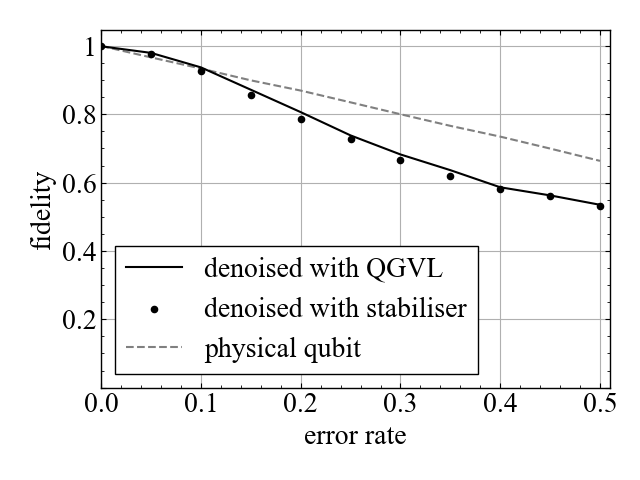}
  \caption{
  Relationship between fidelity and depolarisation error rate for quantum states corrected by the 9-qubit QGVL network trained with the Shor code (solid line), fidelity achieved via algorithmic quantum error correction using the Shor stabiliser code (black dots), and fidelity for physical single-qubit states under the same conditions (dotted line). For both QGVL and the stabiliser code, input states are prepared by encoding the quantum states with the Shor code and introducing depolarisation errors with error rate $p$ to each qubit.
  }
  \label{fig:919D}
\end{figure}

\begin{figure}[!tb]
  \centering
  \includegraphics[width=8cm]{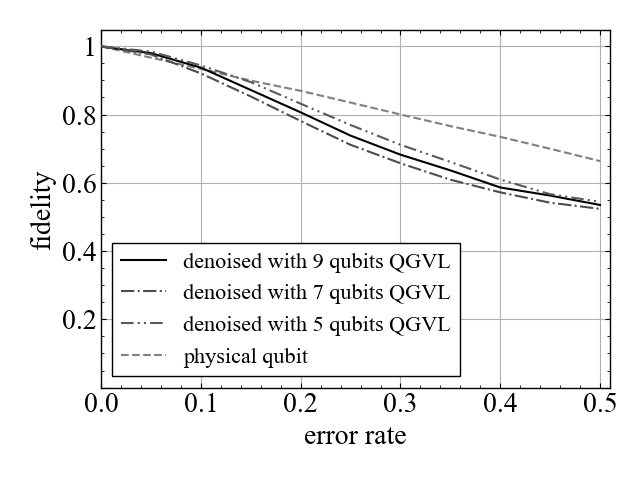}
  \caption{
  Fidelity comparison for QGVL networks with the 5-qubit error correction code, Steane code, and Shor code. Results presented in Figs.~\ref{fig:515D},~\ref{fig:717D}, and~\ref{fig:919D} are compared against fidelities for physical qubits.
  }
  \label{fig:579}
\end{figure}

Figure~\ref{fig:919D} shows the fidelity obtained by 9-qubit QGVL after learning with the Shor code. Unlike the five- or 7-qubit (Steane) codes, fidelity with the Shor code slightly exceeded that obtained by the stabiliser code. This is attributable to the Shor code's degenerate nature, allowing the network to learn and correct a greater variety of error patterns. Additional comparisons among the 5-qubit code, Steane code, and Shor code are shown in Fig.~\ref{fig:579}, where the 5-qubit code consistently yields the highest fidelity, followed by the Shor code and then the Steane code, regardless of error rate. The relative improvement is due to the balance between the number of newly possible errors introduced by additional qubits and the capacity for error correction.

Overall, the Shor code demonstrated higher error correction performance than the Steane code owing to its degenerate structure, while the Steane code is not considered optimal for correcting depolarisation errors with equally probable $X$, $Y$, and $Z$ errors.

\section{Exploring encoding method}

To this point, we have assessed the learning capabilities of QGVL for quantum error correction (QEC) using predefined encoding schemes. As a result, we identified optimal error correction strategies within logical code spaces spanned by, for example, repetition codes and 5-qubit error correction codes. Nevertheless, QEC restricted to predefined coding methods cannot fully leverage the potential to tailor error correction strategies to the specific characteristics of quantum hardware. Therefore, in the present work, we aimed to discover novel logical encoding methods capable of preserving quantum information under previously undefined quantum errors, thereby enabling more versatile QEC strategies.

To accomplish this, we trained two distinct network architectures. The first network is focused on identifying logical encodings that are robust against quantum noise, while the second network performs error correction tailored to the discovered encoding.

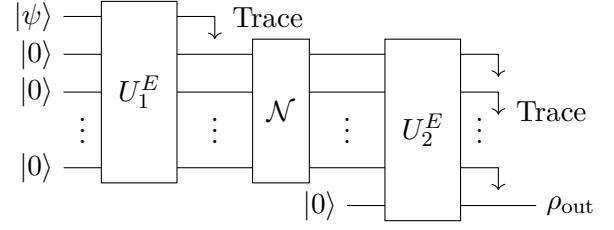
\begin{figure}[!tb]
\centering
\begin{tikzpicture}
\draw(0.5,10)node[left]{$\ket{\psi}$}--(1,10);
\draw(0.5,9.5)node[left]{$\ket{0}$}--(1,9.5);
\draw(0.5,9)node[left]{$\ket{0}$}--(1,9);
\draw(0.75,8.6)node{$\vdots$};
\draw(0.5,8)node[left]{$\ket{0}$}--(1,8);
\draw(1.5,9)node{$U_1^E$};
\draw(1,7.8)--(2,7.8)--(2,10.2)--(1,10.2)--cycle;
\draw(2.6,10)node[right]{$\mathrm{Trace}$};
\draw[->](2,10)--(2.5,10)->(2.5,9.7);
\draw(2,9.5)--(3,9.5);
\draw(2,9)--(3,9);
\draw(2.5,8.6)node{$\vdots$};
\draw(2,8)--(3,8);
\draw(3.375,8.75)node{$\mathcal{N}$};
\draw(3,7.8)--(3.75,7.8)--(3.75,9.7)--(3,9.7)--cycle;
\draw(3.75,9.5)--(4.75,9.5);
\draw(3.75,9)--(4.75,9);
\draw(4.25,8.6)node{$\vdots$};
\draw(3.75,8)--(4.75,8);
\draw(5.25,8.5)node{$U_2^E$};
\draw(4.75,7.3)--(5.75,7.3)--(5.75,9.7)--(4.75,9.7)--cycle;
\draw(6.35,8.75)node[right]{$\mathrm{Trace}$};
\draw(4.25,7.5)node[left]{$\ket{0}$}--(4.75,7.5);
\draw[->](5.75,9.5)--(6.25,9.5)->(6.25,9.2);
\draw[->](5.75,9)--(6.25,9)->(6.25,8.7);
\draw(6,8.6)node{$\vdots$};
\draw[->](5.75,8)--(6.25,8)->(6.25,7.7);
\draw(5.75,7.5)--(6.75,7.5)node[right]{$\rho_\mathrm{out}$};
\end{tikzpicture}
\caption{
  Quantum neural network for learning the encoding unitary matrix $U_1^E$ that maps 1-qubit quantum information $\ket{\psi}$ into a higher-dimensional code space. This network adopts an inverted QEC architecture, acting as an encoder for QEC. Specifically, the Kronecker product of the 1-qubit input state $\ket{\psi}$ with the necessary number of ancilla qubits is computed. The encoded quantum state is then represented as the partial trace over the initial 1-qubit. The noise channel $\mathcal{N}$ subsequently acts on the encoded state, introducing errors to be learned. Finally, the Kronecker product with an additional 1-qubit initial state $\ket{0}$ and a subsequent partial trace yield the output 1-qubit quantum state, $\rho_{\mathrm{out}}$. By optimizing the fidelity between $\rho_{\mathrm{out}}$ and $\ket{\psi}$, the network learns noise-tolerant encoding unitary matrices.
}
\label{fig:explore}
\end{figure}

The exploratory network is a QGVL model with the architecture illustrated in Fig.~\ref{fig:explore}. Unlike previous networks, which adopted an $m$-1-$m$ qubit structure—where the number of qubits in the input and output layers exceeds that in the hidden layer—this network exchanges the roles of encoder and decoder, resulting in a 1-$m$-1 qubit configuration. The noise channel $\mathcal{N}$ is positioned between two unitary operations, which serve as trainable parameters.

The network first encodes a single-qubit state vector into multiple qubits using the initial unitary operation, $U_1^E$, given a predefined or trainable encoding method. The resulting encoded qubits are then exposed to targeted noise processes via the noise channel. The second unitary operation, $U_2^E$, subsequently removes the noise in an attempt to recover the original single-qubit quantum information. By training the network to maximise the similarity between output and input states, it is possible to identify encoding methods enabling reversible decoding of quantum information, even in the presence of noise.

The second network is a QGVL model depicted in Fig.~\ref{fig:explore_qec}. This network learns QEC strategies tailored to the encoding discovered by the first network, differing from previous QGVL architectures only in the use of the newly learned encoding. The process of initial encoding and noise application follows:

\begin{figure*}[!tb]
\centering
\tikzstyle{mynode}=[thick,draw=black!60,fill=black!20,circle,minimum size=5]
\begin{tikzpicture}
  \readlist\Nnod{3,2,1,2,3}
  \foreachitem \N \in \Nnod{
    \foreach \i [evaluate={\x=4.125+\Ncnt*2/3; \y=8.75+(\N/2-\i+0.5)*2/3; \prev=int(\Ncnt-1);}] in {1,...,\N}{
      \node[mynode] (N\Ncnt-\i) at (\x,\y) {};
      \ifnum\Ncnt>1
        \foreach \j in {1,...,\Nnod[\prev]}{
          \draw[thick] (N\prev-\j) -- (N\Ncnt-\i);
        }
      \fi
    }
  }
\draw(0.5,10)node[left]{$\ket{\psi}$}--(1,10);
\draw(0.5,9.5)node[left]{$\ket{0}$}--(1,9.5);
\draw(0.5,9)node[left]{$\ket{0}$}--(1,9);
\draw(0.75,8.6)node{$\vdots$};
\draw(0.5,8)node[left]{$\ket{0}$}--(1,8);
\draw(1.5,9)node{$U_1^E$};
\draw(1,7.8)--(2,7.8)--(2,10.2)--(1,10.2)--cycle;
\draw(2.6,10)node[right]{$\mathrm{Trace}$};
\draw[->](2,10)--(2.5,10)->(2.5,9.7);
\draw(2,9.5)--(3,9.5);
\draw(2,9)--(3,9);
\draw(2.5,8.6)node{$\vdots$};
\draw(2,8)--(3,8);
\draw(3.375,8.75)node{$\mathcal{N}$};
\draw(3,7.8)--(3.75,7.8)--(3.75,9.7)--(3,9.7)--cycle;
\draw(3.75,9.5)--(4.25,9.5);
\draw(3.75,9)--(4.25,9);
\draw(4,8.6)node{$\vdots$};
\draw(3.75,8)--(4.25,8);
\draw(6.125,9.4)node{$\mathcal{Q}$};
\draw(4.25,7.8)--(8,7.8)--(8,9.7)--(4.25,9.7)--cycle;
\draw(8,9.5)--(9,9.5);
\draw(8,9)--(9,9);
\draw(8.25,8.6)node{$\vdots$};
\draw(8,8)--(9,8);
\draw [decorate,decoration={brace,amplitude=5pt,mirror, raise=3pt},yshift=0pt](9,8)--(9,9.5)node[black,midway,xshift=0.7cm]{$\rho_\mathrm{out}$};
\end{tikzpicture}
\caption{
  Quantum neural network for QEC incorporating a trained encoding unitary matrix $U_1^E$. In the first stage, as shown in Fig.~\ref{fig:explore}, a single-qubit quantum state $\ket{\psi}$ is encoded into a multi-qubit state. The encoded state is then subjected to noise via the channel $\mathcal{N}$. The quantum neural network $\mathcal{Q}$ is subsequently trained to correct errors, similarly to the procedure in previous experiments. Once the effective encoding method is identified in the noisy channel, the network can learn to remove the introduced errors from the encoded information.
}
\label{fig:explore_qec}
\end{figure*}
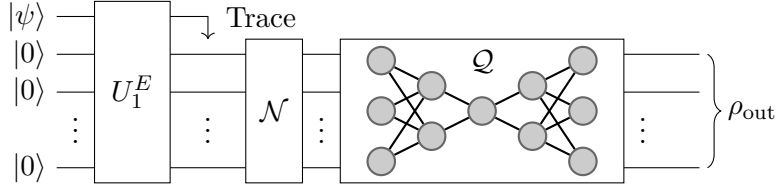

\begin{equation}
    \label{eq:explore_encoding}
    \rho_\mathrm{in} = \mathcal{N}(\underset{\psi}{\mathrm{Tr}}(U_1^E(\ket{\psi}\bra{\psi}\otimes\ket{0}\bra{0}^{\otimes n})U_1^{E\dag}))
\end{equation}

We trained the network using QGVL, employing the encoded quantum state combined with noise as the input state and the corresponding noise-free quantum state as the target state. The original quantum state can be recovered from the encoded quantum state using the first decoding unitary matrix, $U_2^E$.

During training of the first network, $10^4$ randomly generated quantum states were used as both input and target states. As the network introduces noise internally, the unitary matrices are optimised using these identical input and target sets. Subsequently, after generating another set of $10^4$ random quantum state vectors, the second network encodes the quantum states using the unitary matrix learned by the first network.

\begin{figure}[!tb]
  \centering
  \includegraphics[width=8cm]{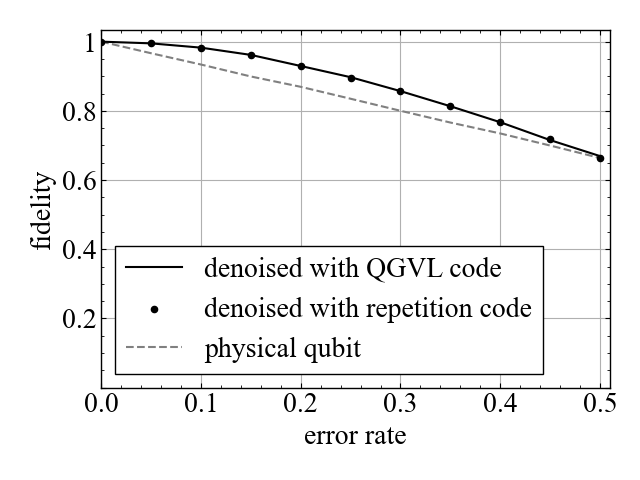}
  \caption{
    Fidelity attained by the 3-qubit QGVL while exploring new encoding methods for bit-flip error correction. The size of the dataset for both training and testing is $10^4$. For comparison, results are plotted alongside those for error correction using the repetition code and corresponding stabiliser code shown in Fig.~\ref{fig:313X}, as well as the fidelity of the uncorrected physical code.
  }
  \label{fig:explore_3qubits}
\end{figure}

The results are presented in Fig.~\ref{fig:explore_3qubits}. We achieved fidelity levels comparable to those obtained by error correction with the assigned encoding methods using the stabiliser code, as well as with QEC performed by QGVL.

\begin{figure}[!tb]
  \centering
  \includegraphics[width=8cm]{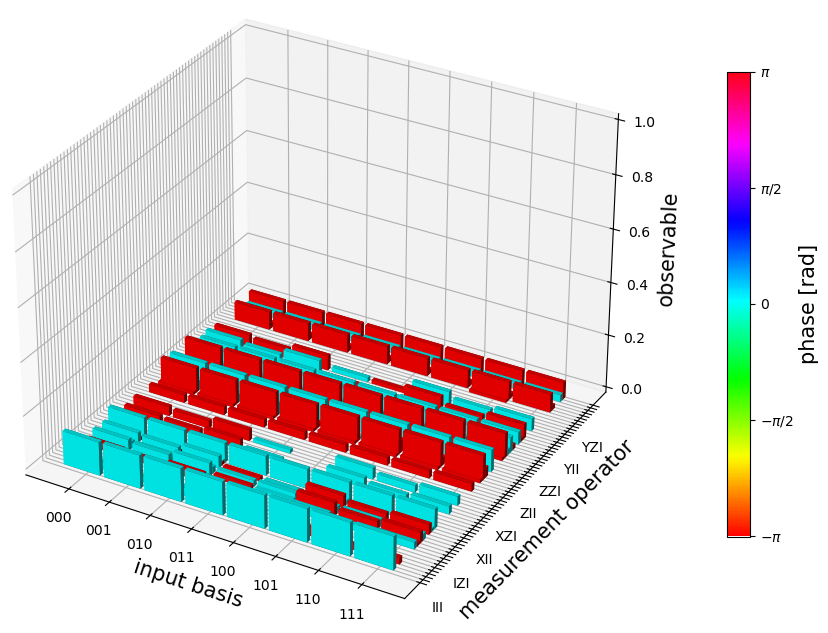}
  \caption{
    QPT of error correction using QEC trained via QGVL. Quantum operations are visualised with the input quantum state $U_1^E (\ket{\psi})$ and the output $\mathcal{Q}(U_1^E (\ket{\psi}))$ as generated by the quantum neural network.
  }
  \label{fig:explore_tomography}
\end{figure}

It should be noted that the code space onto which the original quantum state is projected is a novel code space discovered by the current model. To visualise this, we performed quantum process tomography (QPT, see Appendix~\ref{ap:qpt} for details) for the encoder obtained through training. The results, shown in Fig.~\ref{fig:explore_tomography}, demonstrate that, in contrast to previous training procedures using predefined encodings, the quantum states are mapped to an entirely different code space.

\section{Denoising with noisy channel}
\label{sec:noisy}
Up to this point, our experiments have operated under the idealised assumption of a noise-free quantum neural network (QNN). However, in practice, quantum states may be subject to unexpected errors induced either during idle periods or through gate operations, typically arising from interactions with the environment. In this section, we evaluate quantum error correction (QEC) performance in the presence of internal noise within QNNs such as QGVL, thereby conducting experiments under more realistic conditions.

Our investigations focus on the correction of bit-flip errors. We first examine error correction using a 3-qubit QGVL network, which represents the simplest possible architecture for this purpose. The correction capability of the QGVL network is assessed by comparing its outputs to those of a conventional QAE. Subsequently, we expand the analysis to networks with increased width, specifically 5-qubit and 7-qubit architectures. Finally, we implement more redundant structures, including a 5-3-5 qubit network and a 7-1-7 qubit network, in order to assess the impact of wider network configurations on error correction performance.

\subsection{3-qubit error correction code}
\label{sec:noisy_3-qubit}
We investigated the correction of bit-flip errors in a 3-qubit QGVL network subject to internal noise. The experimental setup was identical to that described in Section~\ref{sec:denoised_bitflip}. For encoding, the 3-qubit repetition code was employed, as defined in Eq.~\eqref{eq:3rep}, with the corresponding stabiliser generators specified in Eq.~\eqref{eq:3stab}. Test data were prepared as described in Section~\ref{sec:denoised_bitflip}.

We now elaborate on the application of network noise, which is a critical aspect of the present experiment. Because the probability of noise generation within the network depends on the characteristics of the quantum device and its operational environment, precise estimation of its fluctuation is generally infeasible. Thus, in this study, we assume that noise affecting different quantum states is equally probable.

Quantum information propagating through the forward layers of the network is successively transformed by interlayer mappings. In networks with two such mappings, three distinct quantum states exist by virtue of the transformation between layers. Moreover, since the input quantum states already contain error due to the generation of training data, the impact of internal network noise in our experiment is represented by applying the noise channel $\mathcal{N}$, with generation probability $p_n$, to quantum states following each interlayer mapping.

When evaluating QEC performance for QNNs operating in noisy conditions, it is necessary to take into account additional noise introduced by the network itself, beyond the input errors. This effect is determined by both the network noise probability $p_n$ and the input error rate $p$. We assessed the noise tolerance of QEC by identifying risk-benefit pairs $(p, p_n)$ that balance these factors.

\begin{figure*}[!tb]
\centering
    \begin{subfigure}{0.49\linewidth}
        \centering
        \includegraphics[width=8cm]{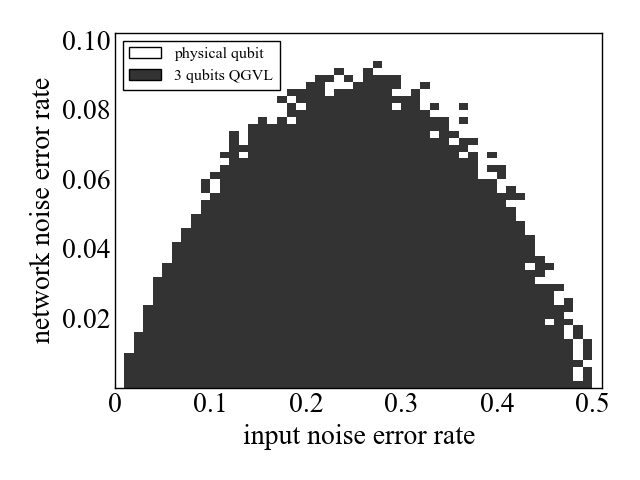}
        \caption{QGVL
        }
        \label{fig:noisy_qgvl}
    \end{subfigure}
    \begin{subfigure}{0.49\linewidth}
        \centering
        \includegraphics[width=8cm]{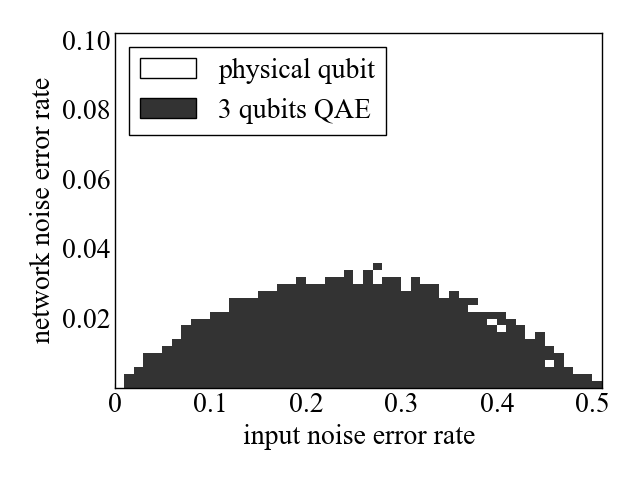}
        \caption{QAE
        }
        \label{fig:noisy_qae}
    \end{subfigure}
    \caption{
        Ranges of input noise error rate $p$ and network noise error rate $p_n$ in which the fidelity achieved by QGVL (a) and QAE (b) exceeds that of the physical qubit. Outcomes for network noise error rates $p_n$ ranging from 0.00 to 0.10 are displayed. Both networks employed the 3-qubit repetition code for encoding. The domain in which QGVL outperforms the physical qubit is 2.77 times larger than the corresponding domain for QAE, and at the highest $p_n$ value, this ratio is 2.70.
        }
\label{fig:noisy_qgvl_qae}
\end{figure*}

Figures~\ref{fig:noisy_qgvl_qae}(a) and~(b) present comparison results for the ranges of $p$ and $p_n$ for QGVL and QAE, along with the corresponding physical qubit results. The QGVL network attained superior fidelity across a broader range of input error rates and network noise rates, indicating greater tolerance to network noise compared to the QAE.

\begin{table*}[!tb]
\centering
\begin{tabular}{lrrrrrr} \hline
error rate & 0.0 & 0.1 & 0.2 & 0.3 & 0.4 & 0.5\\ \hline
physical & \bf{1.00} & 0.933 & 0.866 & 0.799 & 0.734 & \bf{0.666} \\
QGVL & 0.982 & \bf{0.966} & \bf{0.916} & \bf{0.838} & \bf{0.754} & 0.662 \\
QAE & 0.955 & 0.934 & 0.891 & 0.815 & 0.736 & 0.638\\ \hline
\end{tabular}
\caption{
    Fidelity scores for QGVL and QAE with network noise $p_n = 0.02$. Both QGVL and QAE outperform the physical qubit across the ranges illustrated in Fig.~\ref{fig:noisy_qgvl_qae}.
}
\end{table*}

\subsection{Noisy channel with more qubits}
\label{sec:noisy_nqubits}
It is essential to examine experimentally the impact of increasing the number of qubits within a QNN subject to internal noise. Under noiseless conditions, expanding the code length by using more qubits in QGVL networks leads to enhanced error correction performance. However, when internal network noise is present, an increased number of qubits within the network also means more qubits are exposed to noise, resulting in a decline in overall fidelity.

We assessed the noise tolerance of QGVL networks as the number of input qubits was increased from three to five and seven. For training, the repetition code was utilised as in the noiseless case (see Section~\ref{sec:denoised_bitflip_3qubits}). Performance evaluation employed the risk-profit analysis introduced previously in Section~\ref{sec:noisy_3-qubit}.

\begin{figure*}[!tb]
\centering
    \begin{subfigure}{0.32\linewidth}
        \centering
        \includegraphics[width=5.3cm]{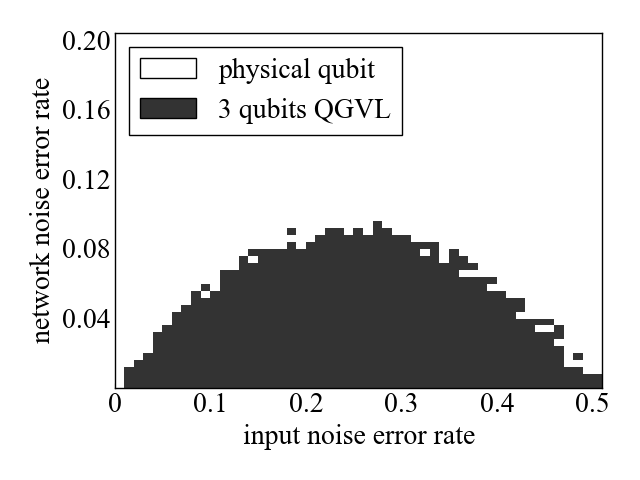}
        \caption{3-qubit
        }
        \label{fig:noisy_X3}
    \end{subfigure}
    \begin{subfigure}{0.32\linewidth}
        \centering
        \includegraphics[width=5.3cm]{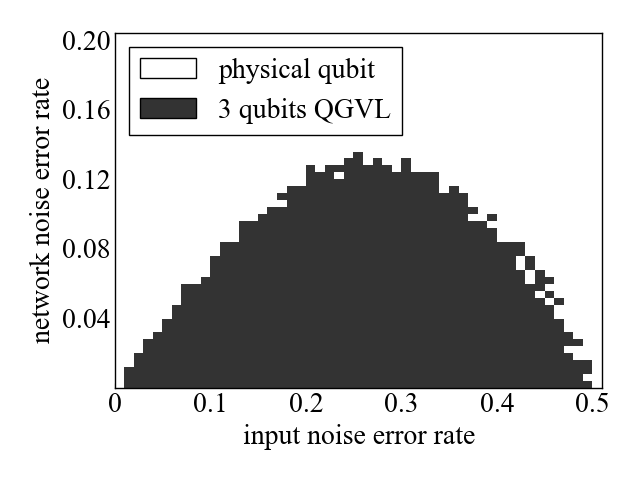}
        \caption{5-qubit
        }
        \label{fig:noisy_X5}
    \end{subfigure}
    \begin{subfigure}{0.32\linewidth}
        \centering
        \includegraphics[width=5.3cm]{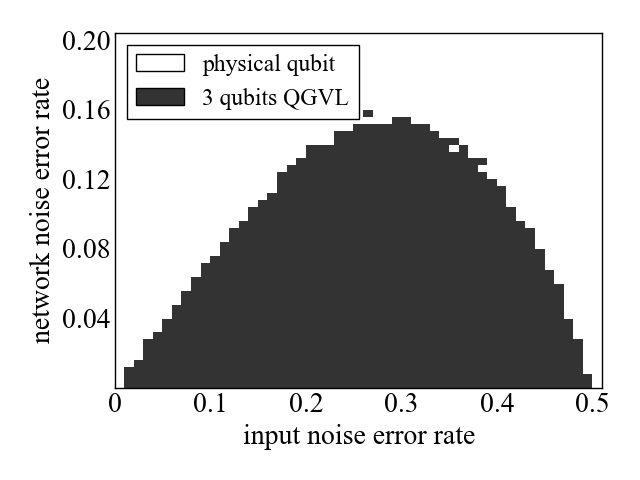}
        \caption{7-qubit
        }
        \label{fig:noisy_X7}
    \end{subfigure}
    \caption{
        Range of input noise error rate $p$ and network noise error rate $p_n$ for which the fidelity of 3-, 5-, and 7-qubit QGVL networks under internal noise exceeds the fidelity of a single physical qubit. The range of $p_n$ spans from 0.00 to 0.20. The maximum probability of network noise generation is 1.41 times higher for the 5-qubit network and 1.62 times higher for the 7-qubit network compared to the 3-qubit case. Likewise, the area under the curve increases by factors of 1.40 and 1.63 for the five- and 7-qubit networks, respectively, compared to the 3-qubit network.
        }
    \label{fig:noisy_X357}
\end{figure*}

Figure~\ref{fig:noisy_X357} shows a comparison of the fidelities achieved by QGVL networks of three, five, and seven qubits under noisy conditions, relative to that of a single physical qubit. As the number of qubits in the QGVL network increases, error correction performance improves, evidenced by a broader region in which fidelity exceeds that of the physical qubit.

\subsection{Multiple neurons in hidden layer}
\label{sec:noisy_hidden}

Until now, the hidden layer in our networks has consisted of a single qubit. In the presence of internal network noise, a hidden layer composed of only 1-qubit, as in the 3-1-3 architecture, carries the risk of complete information loss if that qubit is corrupted. To address this vulnerability, we introduced multiple qubits into the hidden layer, which ensures that not all quantum information is lost when an error occurs. Specifically, we implemented a 5-3-5 network, where the hidden layer contains three qubits and both the input and output layers each contain five qubits. All layers are encoded using the 5-qubit error correction code. This modification is expected to improve the break-even error rate for effective error correction. Increasing the number of qubits in the hidden layer does raise the computational cost for training and operation because the size of the unitary matrices depends on both the input/output and hidden layer dimensions. However, the scalability achieved by QGVL allows these experiments to be conducted with reasonable computational and time resources.

In the experiment, we used a 5-3-5 QNN with 5-qubit input and output layers and a 3-qubit hidden layer. The training data and stabiliser code were the same as those used in Section~\ref{ch:5-qubit}. Expanding the width of the hidden layer required increasing the size of the unitary matrices, from six qubits in the 5-1-5 structure to eight qubits in the 5-3-5 network. To evaluate effectiveness, we compared the fidelity of the 5-3-5 QGVL to that of the 7-1-7 QGVL, which also uses eight total qubits in the input and hidden layers, under conditions of internal network noise. For this comparison, we plotted the pairs of input error rate and network noise error rate that yielded the highest fidelities for physical qubits, 5-3-5 QGVL, and 7-1-7 QGVL.

\begin{figure}[!tb]
\centering
\includegraphics[width=8cm]{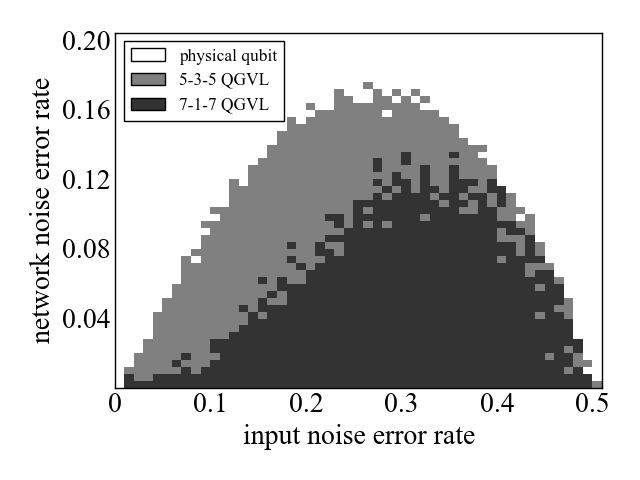}
\caption{
    Error correction capabilities of the 5-3-5 QGVL with multiple qubits in the hidden layer compared to the 7-1-7 QGVL. The maximal fidelity attained for each combination of input error rate $p$ and network noise rate $p_n$ is shown: white for the physical qubit, light grey for the 5-3-5 QGVL, and dark grey for the 7-1-7 QGVL. Encoding was performed using the five- and 7-qubit repetition codes, respectively.
    }
\label{fig:fid_535}
\end{figure}

The results are shown in Fig.~\ref{fig:fid_535}. Under noise-free conditions, the 5-3-5 QGVL achieved fidelity similar to the 5-1-5 QGVL. However, in the presence of internal network noise, the 5-3-5 QGVL outperformed the 7-1-7 QGVL, even though the latter showed superior performance in noise-free situations. Notably, the 5-3-5 QGVL provided better fidelity in regions with higher network noise and lower input error rates, thereby expanding the parameter space where its fidelity surpassed that of the physical qubit.

\begin{figure}[!tb]
\centering
\begin{tikzpicture}
\draw [decorate,decoration={brace,amplitude=5pt, raise=3pt},yshift=0pt](0,8)--(0,10)node[black,midway,xshift=-0.7cm]{$\rho_\mathrm{in}$};
\draw(0,10)--(1,10);
\draw(0,9.5)--(1,9.5);
\draw(0,9)--(1,9);
\draw(0,8.5)--(1,8.5);
\draw(0,8)--(1,8);
\draw(0.5,7.5)node[left]{$\ket{0}$}--(1,7.5);
\draw(0.5,7)node[left]{$\ket{0}$}--(1,7);
\draw(0.5,6.5)node[left]{$\ket{0}$}--(1,6.5);
\draw(1.5,8.25)node{$U_1$};
\draw(1,6.3)--(2,6.3)--(2,10.2)--(1,10.2)--cycle;
\draw(2.6,9)node[right]{$\mathrm{Trace}$};
\draw[->](2,10)--(2.5,10)->(2.5,9.7);
\draw[->](2,9.5)--(2.5,9.5)->(2.5,9.2);
\draw[->](2,9)--(2.5,9)->(2.5,8.7);
\draw[->](2,8.5)--(2.5,8.5)->(2.5,8.2);
\draw[->](2,8)--(2.5,8)->(2.5,7.7);
\draw(2,7.5)--(3,7.5);
\draw(2,7)--(3,7);
\draw(2,6.5)--(3,6.5);
\draw(2.5,6)node[left]{$\ket{0}$}--(3,6);
\draw(2.5,5.5)node[left]{$\ket{0}$}--(3,5.5);
\draw(2.5,5)node[left]{$\ket{0}$}--(3,5);
\draw(2.5,4.5)node[left]{$\ket{0}$}--(3,4.5);
\draw(2.5,4)node[left]{$\ket{0}$}--(3,4);
\draw(3.5,5.75)node{$U_2$};
\draw(3,3.8)--(4,3.8)--(4,7.7)--(3,7.7)--cycle;
\draw(4.6,7)node[right]{$\mathrm{Trace}$};
\draw[->](4,7.5)--(4.5,7.5)->(4.5,7.2);
\draw[->](4,7)--(4.5,7)->(4.5,6.7);
\draw[->](4,6.5)--(4.5,6.5)->(4.5,6.2);
\draw [decorate,decoration={brace,amplitude=5pt,mirror, raise=3pt},yshift=0pt](5,4)--(5,6)node[black,midway,xshift=0.7cm]{$\rho_\mathrm{out}$};
\draw(4,6)--(5,6);
\draw(4,5.5)--(5,5.5);
\draw(4,5)--(5,5);
\draw(4,4.5)--(5,4.5);
\draw(4,4)--(5,4);
\end{tikzpicture}
\caption{
    Architecture of the 5-3-5 QGVL network. The use of multiple qubits in the hidden layer enhances network fault tolerance by protecting against internal network noise that may arise during computation. In the 5-1-5 structure, information is irretrievably lost if the single intermediate qubit is corrupted by noise after the $U_1$ operation. In contrast, the 5-3-5 configuration introduces redundancy not only in the input layer but also in the hidden layer, enabling error correction via the $U_2$ operation in the decoder, even if one of the three intermediate qubits is affected by noise.
    }
\label{fig:535}
\end{figure}
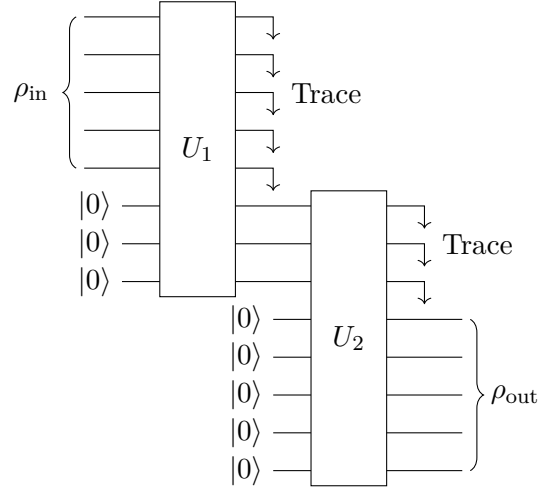
\section{Learning performance}
\label{sec08}

Finally, we investigated enhancements in learning performance enabled by the simplified QGVL architecture. Quantum error correction (QEC) utilising quantum machine learning methods, such as variational quantum algorithms (VQAs), has often been criticised for its vulnerability to the barren plateau phenomenon which is characterised by the vanishing of gradients during training. In the QAE framework, because the unitary matrix parameters are randomly initialised, the network’s ability to avoid barren plateaus critically depends on initial conditions. As a result, achieving optimal quantum error correction typically requires multiple random restarts. Moreover, as network depth and width increase, the intractability of training escalates, undermining scalability for larger or more complex tasks.

Our proposed QGVL method achieves a significant reduction in the number of unitary matrices (i.e., network parameters), thereby mitigating the risk of barren plateaus caused by excessive parameterization. In this section, we systematically assess the training capability and scalability of QGVL by comparing training time and convergence probability to those of the QAE.

\begin{equation}
    \begin{split}
    \mathcal{N}(\rho)=&(1-p)\rho+\frac{p}{3}(X_1\rho X_1+X_2\rho X_2+X_3\rho X_3)
    \end{split}
\end{equation}

During training, pairs $(\mathcal{N}(\psi_L),\psi_L)$, where $\psi_L$ denotes the logical code and $\mathcal{N}$ applies the noise channel, were used as input and target states respectively.

Using these data, we independently trained QGVL and QAE networks. In order to compare the convergence behavior and required training time, we established two distinct criteria for terminating the training process.

\begin{equation}
    \begin{cases}
        |C_t - C_{t-1}| < 10^{-8} \\
        \displaystyle C_t < \frac{1}{N}\sum_{i=1}^{N}L(\rho_\mathrm{targ}^i, \mathrm{QEC}(\rho_\mathrm{in}^i))
    \end{cases}
    \label{eq:learn-finish}
\end{equation}

Here, $c_t$ denotes the value of the cost function at step $t$, and $\mathrm{QEC}$ represents the quantum error-corrected state obtained using the stabiliser code corresponding to the applied noise. Empirically, when training converges for a given dataset, the resulting fidelity often surpasses that of conventional error correction. Therefore, training is considered complete when parameter updates become sufficiently small after adequate progress has been made.

In our experiments, we also set a maximum number of parameter updates for the network, recording the time taken for training to reach completion. Each network was trained $10^3$ times to facilitate robust comparison of training times and convergence steps.

QGVL showed substantial improvements over QAE in both training success rate and time to convergence. The results are summarised in Table~\ref{table:perform}. For example, QGVL achieved perfect completion in all training runs, improving the training success rate by 12\% for 3-qubit and by 25\% for 5-qubit architectures compared to QAE. These findings suggest that QGVL successfully avoids the barren plateau problem, which has been a significant challenge for QAE.

The mean training time was reduced by 80\% in the 3-qubit case and by 96\% in the 5-qubit case, indicating that the reduction in trainable parameters leads to much more efficient training. Across all cases, the standard deviation in training time was less than 2.4\% of the mean, suggesting that convergence times were tightly concentrated and that local minima were largely avoided.

QGVL also reduced the average number of training steps by 7\% for 3-qubit networks and by 57\% for the more complex 5-qubit networks. Random initialisation facilitated rapid convergence and enabled escape from local optima, despite the possibility of barren plateaus. Differences between minimum and maximum step counts reflect the influence of initialisation; early convergence sometimes resulted from fortuitously chosen initial parameters.

Finally, QGVL reduced the mean learning time per step by 36\% for 3-qubit and by 92\% for 5-qubit networks compared to QAE. This improvement is a direct consequence of reduced training time per epoch due to the streamlined model architecture.

\begin{table*}[ht]
\center
\begin{tabular}{lrrrrr} \hline
\multirow{2}{*}{} & \multicolumn{3}{c}{QGVL} & \multicolumn{2}{c}{QAE} \\
& 3-qubit & 5-qubit & 7-qubit & 3-qubit & 5-qubit\\ \hline
Success rate of training [\%] & $\mathbf{100}$ & $\mathbf{100}$ & 97 & 89 & 80\\
Average of training time [s] & $\mathbf{325}$ & 429 & 1579 & 1629 & 13715\\
Standard deviation of training time [$s^2$] & $\mathbf{54}$ & 111 & 254 & 327 & 3508 \\
Average of steps & $\mathbf{123}$ & 144 & 377 & 133 & 341 \\
Standard deviation of steps & $\mathbf{16.0}$ & 37.8 & 40.9 & 20.7 & 74.6 \\
Maximum number of steps & $\mathbf{169}$ & 261 & 459 & 229 & 497 \\
Minimum number of steps & 74 & $\mathbf{61}$ & 97 & 97 & 196 \\
Learning time per step [s] & $\mathbf{2.63}$ & 2.96 & 4.17 & 12.2 & 40.1 \\\hline
\end{tabular}
\caption{
Comparison of total training time and step count for QGVL and QAE. Training hyperparameters: $10^3$ data points, $10^2$ training runs, and a maximum of 500 steps. The input/output layers comprised 3, 5, and 7 qubits for QGVL and QAE, respectively. See \eqref{eq:learn-finish} for stopping criteria. Only successful training runs are included in the calculations, except for the "Success rate of training."
}
\label{table:perform}
\end{table*}
\section{Discussion and outlook}
\subsection{QGVL's positive effect}
In this study, we proposed quantum error correction (QEC) via QGVL as a quantum machine learning based approach to quantum error correction. Using QGVL, we successfully corrected errors occurring in input quantum information within predefined code spaces. This capability mitigates vulnerability to qubit noise, which is a major challenge in the realisation of quantum computers, and enables the construction of quantum memories for storing quantum information. QGVL employs a global unitary matrix as the trainable parameter in the QEC procedure, thereby reducing the number of required unitary matrices compared with previous methods and consequently shortening training time and lowering computational cost.

Furthermore, this reduction in model complexity allows QGVL to train larger networks. Under noiseless conditions, QGVL achieved shorter training times and faster convergence to a (near) globally optimal solution while maintaining accuracy comparable to existing approaches. Even in the presence of depolarizing errors, QGVL preserved this level of accuracy. These results indicate that QGVL is well suited for implementation on real quantum devices subject to various noise mechanisms.

In the case of noisy networks, QGVL also improved error correction accuracy and training efficiency compared with previous work, primarily because the number of unitary operations in the quantum circuit was reduced, which in turn decreased the number of potential noise-generating operations. When internal network noise was present, the QGVL-based network was able to remove noise from encoded quantum information as long as the error probability remained below a certain threshold. QGVL increased the admissible internal noise threshold, thereby enhancing protection of quantum information. This threshold could be further raised by enlarging the network and introducing multiple qubits in the hidden layer.

Moreover, network enlargement became feasible due to shorter training times and fewer occurrences of convergence to local minima. Under the assumption of negligible internal network noise, the admissible error threshold could be increased by enlarging the network through additional input-layer qubits. In contrast, under more realistic conditions with internal network noise, a higher admissible error threshold was obtained by expanding the network via additional hidden-layer qubits.

In the experiments described above, quantum information was first encoded using a predefined code space. By modifying the network architecture, we then realised an encoding in an optimal code space adapted to the given noise model. This extension not only enables QEC for unknown noise channels but also allows channel-adaptive encoding and QEC that are robust against the types of noise that occur in practical quantum devices. As a consequence, the optimal parameters differ for each encoding, which complicates decoding for an adversary and further enhances robustness.

\subsection{Limitation of current study}
In the present study, larger-qubit networks were successfully trained; however, the number of trainable qubits remains limited. The current upper bound is 9 qubits, because training 11-qubit networks could not be completed within a reasonable time. Consequently, the logical error rate remains higher than that of error correction codes employing larger numbers of physical qubits. Further enlargement of the qubit count will require more efficient training schemes and is therefore left for future work.

Another issue to be addressed in future research is the optimization of complex-valued gradients during training. In this study, the absolute value of each matrix element was used as a secondary momentum term, and optimization was performed on the corresponding real-valued norm without distinguishing between real and imaginary components. This procedure relies on a reduced amount of information, since complex values are compressed into real scalars. Enhancing this optimization method to exploit the full complex structure is expected to enable more efficient training and, in turn, QEC on larger and more complex networks.

\bibliographystyle{quantum}
\bibliography{reference}
\clearpage
\onecolumn
\appendix
\section{Training Network}
\label{ap:train}

In this section, we provide a detailed explanation of network training with explicit equations. \par
The objective of the QGVL network $\mathcal{Q}$ is to output noise-free encoded quantum states from encoded quantum information that has been subjected to a specified noise channel. To this end, training data are constructed from pairs of quantum state vectors. First, a randomly generated quantum state vector $\ket{\psi}$ is transformed into an encoded state $\ket{\psi_L}$ according to the encoding scheme appropriate to each experiment. The target state for training is the density operator
\[
\rho_{\mathrm{targ}} = \ket{\psi_L}\bra{\psi_L}.
\]
The corresponding input training data are obtained by applying the noise channel $\mathcal{N}$ to this density operator, yielding
\[
\rho_{\mathrm{in}} = \mathcal{N}(\rho_{\mathrm{targ}}).
\]

To maximise the fidelity between the network output $\rho_{\mathrm{out}} = \mathcal{Q}(\mathcal{N}(\rho_{\mathrm{in}}))$ and the target state $\rho_{\mathrm{targ}}$, we minimise the cost function in Eq.~\eqref{eq:qgvl_cost}, defined as the one-minus-average fidelity:

\begin{equation}
\label{eq:qgvl_cost}
    C = 1 - \frac{1}{N} \sum_{i=1}^{N}\bra{\rho_{\mathrm{targ}}^i}\rho_{\mathrm{out}}^i\ket{\rho_{\mathrm{targ}}^i}
\end{equation}

In the actual training procedure, the unitary matrix in each layer is treated as a trainable parameter and is updated iteratively. Using stochastic gradient descent (SGD) as the underlying optimization method, the unitary matrix $U_k(s+\epsilon)$ at step $s+\epsilon$ is updated according to

\begin{equation}
\label{eq:U_k}
    U_k(s+\epsilon) = e^{i\epsilon K_k(s)}U_k(s)
\end{equation}

so that unitarity of $U_k$ is preserved at each step. \par

The update operator $K_k(s)$ is computed in a manner analogous to backpropagation. For a given layer $k$, the input state $\rho_{\mathrm{in}}^i$ of the $i$-th training sample (out of a total of $N$ samples) is propagated forward through the network up to the unitary $U_k$, while the corresponding target state $\phi_{\mathrm{targ}}^i$ is propagated backward from the output side. The commutator $M_k$ serves as an error operator between these forward- and backward-propagated quantum states:

\begin{equation}
    \begin{split}
    M_k(s,i) = [U_k(s)\ldots U_2(s)\left(\rho_{\mathrm{in}}^i \otimes \ket{0\ldots 0}\bra{0\ldots 0}_{\rm{hidden}}\right)U_2(s)^\dag \ldots U_k(s)^\dag, \\
    U_{k+1}(s)^\dag \ldots U_{out}(s)^\dag \left(\mathbf{1}_{\rm{hidden}} \otimes \ket{\phi_{\rm{targ}}^i}\bra{\phi_{\rm{targ}}^i}\right)U_{\rm{out}}(s)\ldots U_{k+1}(s)]
    \end{split}
    \label{eq:qgvl_M}
\end{equation}

The commutators $M_k(s,i)$ are evaluated for all $N$ training samples, since each commutator encodes the error contribution from the $i$-th data point at layer $k$. The update operator $K_k(s)$ is then obtained as the averaged commutator:

\begin{equation}
    \label{Kks}
    K_k(s) = i \frac{\rm{dim}(U_k)}{2N} \sum_{i=1}^N{M_k(s,i)}
\end{equation}

Finally, the unitary matrices are updated using Eqs.~\eqref{eq:U_k} and \eqref{Kks}. Compared with QAE, QGVL requires significantly fewer unitary matrices. Consequently, the number of matrices $U_k$, update operators $K_k$, and commutators $M_k$ is reduced, which leads to an expected improvement in computational cost. \par

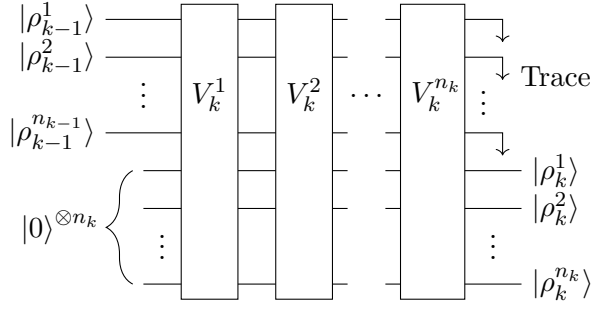
\begin{figure}[!tb]
\centering
\begin{tikzpicture}
\draw(0,10)node[left]{$\ket{\rho_{k-1}^1}$}--(1,10);
\draw(0,9.5)node[left]{$\ket{\rho_{k-1}^2}$}--(1,9.5);
\draw(0.5,9.1)node{$\vdots$};
\draw(0,8.5)node[left]{$\ket{\rho_{k-1}^{n_{k-1}}}$}--(1,8.5);
\draw(1.375,9)node{$V_k^1$};
\draw(1,6.3)--(1.75,6.3)--(1.75,10.2)--(1,10.2)--cycle;
\draw(2.625,9)node{$V_k^2$};
\draw(2.25,6.3)--(3,6.3)--(3,10.2)--(2.25,10.2)--cycle;
\draw(4.375,9)node{$V_k^{n_k}$};
\draw(3.9,6.3)--(4.75,6.3)--(4.75,10.2)--(3.9,10.2)--cycle;
\draw(1.75,10)--(2.25,10);
\draw(1.75,9.5)--(2.25,9.5);
\draw(1.75,8.5)--(2.25,8.5);
\draw(3,10)--(3.2,10);
\draw(3,9.5)--(3.2,9.5);
\draw(3,8.5)--(3.2,8.5);
\draw(3.45,9)node{$\hdots$};
\draw(3.7,10)--(3.9,10);
\draw(3.7,9.5)--(3.9,9.5);
\draw(3.7,8.5)--(3.9,8.5);
\draw(5.35,9.25)node[right]{$\mathrm{Trace}$};
\draw[->](4.75,10)--(5.25,10)->(5.25,9.7);
\draw[->](4.75,9.5)--(5.25,9.5)->(5.25,9.2);
\draw(5,9)node{$\vdots$};
\draw[->](4.75,8.5)--(5.25,8.5)->(5.25,8.2);
\draw [decorate,decoration={brace,amplitude=10pt, raise=4pt},yshift=0pt](0.5,6.5)--(0.5,8.0)node[black,midway,xshift=-1.1cm]{$\ket{0}^{\otimes n_k}$};
\draw(0.5,8.0)--(1,8.0);
\draw(0.5,7.5)--(1,7.5);
\draw(0.75,7.1)node{$\vdots$};
\draw(0.5,6.5)--(1,6.5);
\draw(1.75,8.0)--(2.25,8.0);
\draw(1.75,7.5)--(2.25,7.5);
\draw(1.75,6.5)--(2.25,6.5);
\draw(3,8)--(3.2,8);
\draw(3,7.5)--(3.2,7.5);
\draw(3,6.5)--(3.2,6.5);
\draw(3.7,8)--(3.9,8);
\draw(3.7,7.5)--(3.9,7.5);
\draw(3.7,6.5)--(3.9,6.5);
\draw(4.75,8.0)--(5.5,8.0)node[right]{$\ket{\rho_k^1}$};
\draw(4.75,7.5)--(5.5,7.5)node[right]{$\ket{\rho_k^2}$};
\draw(5.125,7.1)node{$\vdots$};
\draw(4.75,6.5)--(5.5,6.5)node[right]{$\ket{\rho_k^{n_k}}$};
\end{tikzpicture}
\caption{
    Interlayer mapping operations in a DQNN on a classical simulator. Because strictly local unitary operations, as depicted in Fig.~\ref{fig:dqnn-map}, are difficult to implement efficiently on a classical computer, we instead perform operations using the extended unitary matrices defined in Eq.~\eqref{eq:qae-v}. As a consequence, the size of the unitary matrices in interlayer mappings is the same as in the original DQNN, but $n_k$ such operations are required. The total computational cost is further increased by the overhead of extending and compressing unitary matrices in the classical simulation.
}
\label{fig:qae_detail}
\end{figure}

We evaluated the computational cost under the assumption that multiplication of two $N \times N$ matrices scales as $O(N^3)$. In an $m$-$n$-$m$ QAE, there are $n$ square matrices of dimension $2^{m+1}$ and $m$ square matrices of dimension $2^{n+1}$, whereas QGVL employs only two square matrices of dimension $2^{m+n}$. If one naïvely assumes direct unitary matrix multiplication, the corresponding costs are $O(2^{3(m+1)} n + 2^{3(n+1)} m)$ for QAE and $O(2^{3(m+n)+1})$ for QGVL. This comparison would suggest that QAE is less costly than QGVL, even in the simplest 3-1-3 architecture. However, this conclusion does not hold in practice, because on a classical computer it is not feasible to update only small local blocks within highly sparse unitary matrices without effectively working with the full matrix dimension. The actual computation required for QAE therefore follows the structure illustrated in Fig.~\ref{fig:qae_detail}. Consequently, the effective size of the unitary matrices in QAE becomes comparable to that in QGVL, and because QAE requires $(m+n)$ such operations, its total computational cost is in fact larger than that of QGVL. \par

We now derive the computational cost more precisely. For a unitary matrix $U_k^j$ of size $2^{n_{k-1}+1}$, we first take the Kronecker product with the identity matrix of size $2^{n_k-1}$ to obtain an extended unitary matrix $V_k^j$ of size $2^{n_{k-1}+n_k}$:

\begin{equation}
    \label{eq:qae-v}
    V_k^j = \underset{n_k+1,n_k+j}{\mathrm{SWAP}} (U_k^j \otimes I_{n_k-1})
\end{equation}

The computational cost is then estimated using these extended unitary matrices. The interlayer mapping can be written as

\begin{equation}
    \label{eq:qae-map-v}
    \rho_k = \mathcal{E}_k(\rho_{k-1}) = \underset{k-1}{\mathrm{Tr}}[V_k^{n_k}\ldots V_k^1(\rho_{k-1} \otimes \ket{0}^{\otimes n_k}\bra{0}^{\otimes n_k})V_k^{1\dag} \ldots V_k^{n_k\dag}]
\end{equation}

As Eqs.~\eqref{eq:qae-v} and \eqref{eq:qae-map-v} indicate, the size of the effective unitary matrices in QAE is the same as in QGVL. However, because the number of required applications is $(m+n)$ in QAE, its overall computational cost is evidently higher than that of QGVL. \par

We next examine the difference in computational cost between QAE and QGVL in more detail. When the number of training data is sufficiently large, in addition to the matrix multiplications for each of the $n_k$ unitaries in QAE, the application of SWAP gates must also be taken into account. In both cases, the cost per operation scales as $O\!\bigl((n_{k-1}+n_k)^3\bigr)$, and the total cost per layer becomes proportional to $(2n_k - 1)$. When the number of data is small, further costs from computing matrix exponentials, partial traces, and Kronecker products must also be considered. \par

Convergence to an optimal solution in either QAE or QGVL is not guaranteed to proceed at a uniform rate, since the training dynamics depend sensitively on the cost landscape. As a result, the average total training time can vary substantially with the presence of local minima and with the initial choice of unitary matrices, even though the per-step computational cost scales in proportion to the average training time. \par

\section{Quantum Process Tomography}
\label{ap:qpt}

In this study, various quantum operations, including those implemented by QNN-based QEC, were applied to quantum states. Quantum Process Tomography (QPT) was introduced to characterise and visualise these quantum processes. To clarify QPT, we first review Quantum State Tomography (QST), a closely related method. In general, QST is used to experimentally reconstruct an unknown quantum state, whereas QPT extends this idea to reconstruct the action of an unknown quantum process on quantum states.

For simplicity, consider first a single-qubit state. The operators \(\frac{I}{\sqrt{2}}, \frac{X}{\sqrt{2}}, \frac{Y}{\sqrt{2}}, \frac{Z}{\sqrt{2}}\) form an orthonormal basis of operators with respect to the Hilbert–Schmidt inner product. An unknown single-qubit density matrix \(\rho\) can be written as

\begin{equation}
    \label{eq:qst1}
    \rho = \frac{\mathrm{Tr}(\rho)I+\mathrm{Tr}(X\rho)X+\mathrm{Tr}(Y\rho)Y+\mathrm{Tr}(Z\rho)Z}{2}
\end{equation}

Here, \(\mathrm{Tr}(A\rho)\) denotes the expectation value of the observable \(A\). When the number of measurement repetitions \(m\) is sufficiently large, the sample mean of the measurement outcomes approaches \(\mathrm{Tr}(A\rho)\) by the central limit theorem, with an approximately Gaussian distribution of standard deviation \(\frac{\Delta(A)}{\sqrt{m}}\), where \(\Delta(A)\) is the standard deviation of \(A\). Since \(\Delta(A) \le 1\) for Pauli observables, the statistical error of the estimate is at most \(\frac{1}{\sqrt{m}}\).

We now extend this to an \(n\)-qubit state. Using tensor products of Pauli matrices as before, an unknown \(n\)-qubit density operator \(\rho\) can be expanded as

\begin{equation}
    \label{eq:qst}
    \rho = \sum_{\vec{v}}\frac{\mathrm{Tr}(\sigma_{v_1}\otimes\sigma_{v_2}\otimes\hdots\otimes\sigma_{v_n}\rho)\sigma_{v_1}\otimes\sigma_{v_2}\otimes\hdots\otimes\sigma_{v_n}}{2^n}
\end{equation}

where the sum runs over \(\vec{v} = (v_1,\ldots,v_n)\) with \(v_i \in \{0,1,2,3\}\), and \(\sigma_{0},\sigma_{1},\sigma_{2},\sigma_{3}\) denote \(I,X,Y,Z\), respectively. In this way, all possible tensor-product Pauli operators form a basis, and the state \(\rho\) can be reconstructed from measured expectation values.

On this basis, QPT can be described as follows. Let the Hilbert space have dimension \(d\). Choose a set of \(d^2\) input states \(\ket{\psi_1},\ldots,\ket{\psi_{d^2}}\) such that the projectors \(\ket{\psi_j}\bra{\psi_j}\) span the operator space. For each input state, the corresponding output state \(\mathcal{E}(\ket{\psi_j}\bra{\psi_j})\) is reconstructed by QST. Since \(\mathcal{E}\) is linear, its action on arbitrary states can then be expressed as a linear combination in this operator basis, thereby providing a full characterization or “visualisation” of the quantum process \(\mathcal{E}\). In the present study, Pauli operators were employed as the basis to represent errors, and thus were also used as the operator basis for QPT to give a clear description of the underlying error processes and correction operations.

\end{document}